\documentclass[a4paper,fleqn]{cas-dc}
\usepackage[authoryear]{natbib}

\usepackage{graphics}
\usepackage{graphicx}
\usepackage{grffile} %
\usepackage{graphbox}
\usepackage[export]{adjustbox}
\usepackage{multirow}

\usepackage{hhline} %

\usepackage{caption}
\usepackage{subcaption}
\usepackage{setspace}

\usepackage{tikz}
\usepackage{booktabs}

\usepackage{etoolbox}

\usepackage{collcell}

\usepackage{lastpage}
\usepackage{fancyhdr}
\def\tsc#1{\csdef{#1}{\textsc{\lowercase{#1}}\xspace}}
\tsc{WGM}
\tsc{QE}
\tsc{EP}
\tsc{PMS}
\tsc{BEC}
\tsc{DE}
\begin{document}

\let\WriteBookmarks\relax
\def\floatpagepagefraction{1}
\def\textpagefraction{.001}
\shorttitle{A deep scalable neural architecture for soil properties estimation}
\shortauthors{Piccoli et al.}

\title [mode = title]{A deep scalable neural architecture for soil properties estimation from spectral information}

\author[1,2]{Flavio Piccoli}[type=editor,
                        auid=000,bioid=1,orcid=0000-0001-7432-4284]
\credit{Investigation, Conceptualization, Methodology, Software, Formal analysis, Validation, Visualization, Writing $-$ Review \& Editing, Writing $-$ Original Draft}

\author[2]{Micol Rossini}[type=editor,
                        auid=000,bioid=1,orcid=0000-0002-6052-3140]
\credit{Investigation, Conceptualization, Methodology, Formal analysis, Validation, Visualization, Writing $-$ Review \& Editing, Writing $-$ Original Draft}

\author[2]{Roberto Colombo}[type=editor,
                        auid=000,bioid=1,orcid=0000-0003-3997-0576]
\credit{Investigation, Conceptualization, Methodology, Formal analysis, Validation, Visualization, Writing $-$ Review \& Editing, Writing $-$ Original Draft}

\author[2]{Raimondo Schettini}[type=editor,
                        auid=000,bioid=1,orcid=0000-0001-7461-1451]
\credit{Investigation, Conceptualization, Methodology, Formal analysis, Validation, Visualization, Writing $-$ Review \& Editing, Writing $-$ Original Draft}
                        
\author[2]{Paolo Napoletano}[type=editor,
                        auid=000,bioid=1,orcid=0000-0001-9112-0574]
\credit{Investigation, Conceptualization, Methodology, Software, Formal analysis, Validation, Visualization, Writing $-$ Review \& Editing, Writing $-$ Original Draft}

\address[1]{Istituto Nazionale di Fisica Nucleare}
\address[2]{Universit\`a Milano-Bicocca}

\begin{abstract}
In this paper we propose an adaptive deep neural architecture for the prediction of multiple soil characteristics from the analysis of hyperspectral signatures. 
The proposed method overcomes the limitations of previous methods in the state of art: (i) it allows to predict multiple soil variables at once; (ii) it permits to backtrace the spectral bands that most contribute to the estimation of a given variable; (iii) it is based on a flexible neural architecture capable of automatically adapting to the spectral library under analysis. The proposed architecture is experimented on LUCAS, a large laboratory dataset and on a dataset achieved by simulating PRISMA hyperspectral sensor. 'Results, compared with other state-of-the-art methods confirm the effectiveness of the proposed solution.

\end{abstract}


\maketitle

\section{Introduction and background}

In the last few years, the increasing needs of the population and the decrease of cultivable areas loudly demand the rationalization and efficiency of the farming-related processes. 
To address this need, several techniques have been proposed and fall in what is called Precision Agricolture (PA) \citep{pierce1999aspects}.

PA is an approach to farm management that uses information technology (IT) to ensure that soil and crops receive precisely what they need for optimal health and productivity. The goal of PA is to simultaneously ensure sustainability, environmental protection and profitability.

The continuous monitoring of soil properties plays a key role in PA, allowing a better management of resources \citep{ritter}, lower usage of insecticides \citep{nicolopoulou2016chemical} and higher quality of the harvest \citep{gebbers2010precision}.
This process requires the manual collection of soil samples in the area under investigation through time and the chemical-physical analysis of each acquired sample to determine the monitored soil properties.
Despite this process has been standardized and optimized with the time \citep{toth2013LUCAS}, it is very time\- and energy\- consuming and it requires a special and expansive setup.

The physical link between soil components and the electromagnetic spectrum in the optical region is well known.
This relationship allows the characterization of soil through the analysis of its spectral signature, acquired through a spectrograph, making the need of collecting and analyzing each soil sample no longer necessary (\cite{angelopoulou2019remote}.
Furthermore, the analysis of the spectral signature through machine learning techniques allows the automation of this process and, in combination with remote sensing, enables the possibility to extend the learned knowledge to never-seen areas.

Optical remote sensing is a powerful tool for collecting data at different scale and useful to characterize soil properties \citep{mulder, ben2002quantitative}.
Those sensors can be mounted on drones, unmanned aerial vehicles (UAV), aircrafts and satellites. %
In particular satellite data are very attractive because they can cover large areas with short revisit time, resulting in the availability of multiple images over the same area.
Among the available multispectral satellites, in recent years high quality sensors such as Landsat-8 and Sentinel-2 allowed for the accurate estimation of topsoil variables, in particular soil organic carbon \cite{crucil2019assessing}. An improvement of soil variable prediction is expected using the recently launched and forthcoming hyperspectral sensors as the ASI PRISMA (\cite{galeazzi2008prisma, cogliati2021prisma}), DLR EnMap (\cite{hofer2006enmap})) and ESA CHIME.
However, those sources lack of proximity measures that allow the development of supervised Machine Learning methods for the prediction of soil properties. 
Furthermore, airborne\- and spaceborne\-based acquisitions present geometrical and radiometrical distortion given by the uncertainty on the intrinsic and estrinsic parameters of the sensor with respect to the point under observation and the atmospheric interference which cause a severe or total loss of information in some spectral bands~\citep{angelopoulou}.

To support the development of supervised Machine Learning methods, ~\cite{LUCAS} proposed a dataset for soil characteristics estimation from hyperspectral signal acquired under controlled conditions. Due to its cardinality and fullness of information, this dataset became one of the most important benchmarks for assessing soil characteristics estimation for precision agricolture. For this reason, we align with the state of art and use it to test our method.

The state of art is composed by several methods that focus on different aspects of soil properties estimation, however there is a common thread: the characterization of soil from its spectral reflectance, weather it's acquired from proximity, airborn or spaceborn devices.
Following we will describe in detail the methods that represent the state of art of soil parameter estimation using spectral information and machine learning.

\cite{Forkuor} estimate six soil properties on the West Africa terrain starting from satellite multispectral images. The satellites involved in this study are RapidEye and Landsat. Six soil properties are estimated: cation exchange capacity (CEC), sand, silt, clay, nitrogen (N) and soil organic carbon (SOC). The proposed method uses random forests (RF), support vector machines (SVM)  \citep{vapnik2013nature} and stocastic gradient boosting (SGB). The features used are the values of the multispectral bands themselves.

Similarly, \cite{zhou2021prediction} predict SOC and the carbon-nitrogen ratio from multispectral data, covering the Swizerland territory \citep{panagos2012european}.

\cite{guo2019prediction} extend the survey including vis-NIR spectral information to estimate SOC. In first place, a direct method similar to \cite{zhou2021prediction} is presented. In this context, SOC is estimated directly from the hyperspectral signal. In second place, \cite{guo2019prediction} try to estimate SOC from two intermediate estimated variables: the soil organic matter (SOM) and the soil bulk density (SBD).

Following the work of \cite{guo2019prediction}, \cite{meng} propose a multiscale approach to handle details at different level. Instead of using all hyperspectral bands, they estimate SOC directly from the bands which show the maximum correlation with the target variable. Beside the common estimators (RF, SVM) they also use backpropagation neural networks. To address noise in high frequency bands, a low pass filter is applied. The filter is based on a discrete wavelet transform.

\cite{li2020simultaneous} increase the amount of considered soil properties in the estimation process, including nitrogen and clay. The predictor is a multi-branch neural network which estimates the target variables from the raw spectral signal and a 2D spectrum map obtained through a Fourier transform. In this work, it has also been investigated the impact of three different pre-processing methods on the overall performances. Specifically, the preprocessing investigated are: the Savitzky-Golay (S-G) smoothing algorithm \citep{menon2014robust,dombi2020adaptive}, the Multivariate Scattering Correction (MSC) \citep{heil2021evaluation, mancini2019study} and the centralization methods.

Despite SOC is the most studied soil characteristic, \cite{hu2019quantitative} investigate the soil salinity from multispectral data. The method uses random forests to estimate the salinity.

\cite{Castaldi2016} studied the capability of seven multi- and hyper-spectral satellite imagers to estimate soil variables. Four current and three forthcoming satellite imagers were simulated by resampling the spectral signals of two datasets and by adding decleared and actual noise. Partial least square regression was used to estimate clay, sand, silt and organic carbon content.

The base technologies used by these methods are then support vector machines (SVM), random forests (RF) and boosting regression trees (BRT), in turn applied to different sets of data. As you will see in subsection \ref{sub:best-result}, for a fair comparison we adapted these predictors to the new data and we performed a parameter search to find the most suitable set of parameters.

Methods in the state of the art have some limitations. 
First of all, the majority of them are not designed to estimate multiple variables at the same time.
Second, it is very rare for a method to consider a huge amount of data to train and validate the machine learning model. 
Third, it is rare that a method is designed to highlight the portion of the spectrum that mostly contributed to the prediction of a given soil variable.

In this paper we propose a new method based on Deep Neural Networks for the estimation of soil characteristics that overcomes the limitations mentioned above. In addition, the method leverages a neural architecture search to automatically find the architecture that better explains the data under analysis. The proposed method is experimented on Lucas \cite{LUCAS}, a large publicly available laboratory dataset and on a dataset achieved by simulating the PRISMA hyperspectral sensor (\cite{galeazzi2008prisma, cogliati2021prisma}). Results confirm the effectiveness of the proposed solution with respect to the state of art. Below we summarise the main contributions brought by this work:

\begin{itemize}
    \item an adaptive convolutional neural network to estimate multiple soil properties at once;
    \item investigation about which spectral bands are the most important for the estimation of each variable;
    \item application of the proposed method on simulated PRISMA data
\end{itemize}
The paper is organized as follow: in section \ref{sec:proposed-method} we describe the proposed method; in section \ref{sec:data-composition} we discuss the data employed in the experimentation; in section \ref{sec:experiments} we perform several experiments; in section \ref{sec:conclusions} reports conclusions and future works.

\section{Proposed method} \label{sec:proposed-method}
We propose a flexible system for soil characteristics estimation from hyperspectral signal based on convolutional neural networks. The proposed method is designed in such a way that it allows a fast architecture search and a scalable hyper parameter search. The resolution of the input, the network composition in terms of layer types, dimensionality and quantity, the loss function and the hyperparameters are some of the aspects of neural regression that are strictly dependent from each other and a slight variation of one of these elements affects the others. 

To enable the search of the best-performing setup, we define a complete architecture as a composition of building blocks, which are instances of the same building block with different parameters. In the next section \ref{sub:building-block} we introduce the building block while in subsection \ref{sub:architectural-composition} we describe how to combine them to obtain the final architecture.

\subsection{Building block}
\label{sub:building-block}
Table \ref{tab:building-block-composition} represents the structure of the building block. 
A first group of layers halves the size of the input map and then an optional sequence of $N$ refinement groups analyze the feature map, increasing the feature abstraction and the field of view of the building block.
The block is fully configurable.
The parameters $\psi$ characterizing the structure of the building block are:

\begin{itemize}
\item $CH_{in}$ - the number of input channels
\item $CH_{out}$ - the number of output channels
\item $N$ - the number of refinement blocks
\item $norm$ - the use of batch normalization or not
\item $leak$ - if zero, it uses ReLU otherwise a LeakyReLU with that value
\end{itemize}

The amount of parameters of parameters for the downsampling block are:
\begin{equation}
    |\theta_{dwn}| = 4 \times CH_{in} \times CH_{out} + norm \times 2 \times CH_{out}
\end{equation}

while each refinement block has:

\begin{equation}
    |\theta_{ref}| = 3 \times CH_{out} \times CH_{out} + norm \times 2 \times CH_{out}
\end{equation}

parameters. The total amount of parameters is then:

\begin{equation}
    |\theta_{blk}| = |\theta_{dwn}| + N \times |\theta_{ref}|
\end{equation}

The field of view given by each block is:

\begin{equation}
    FOV = 4 + N \times 3
\end{equation}

\begin{table}
\begin{center}
  \caption{Structure of the building block. It's composed by a first group of blocks that downsample the input map at half of its original resolution and then by N optional refinements. Each group of blocks is composed by a convolution that can be followed by a normalization layer and then by a non-linearity, which in turn can be a LeakyReLU or a ReLU.}
  \label{tab:building-block-composition}
  \resizebox{.5\textwidth}{!}{ %
  \begin{tabular}{llc}
    \toprule
    Stage & Operation & Output size \\ \midrule
    Pre-processing & Input & $CH_{in} \times R$ \\
    \addlinespace
    Downsampling & Conv $4$ stride $2$ & \multirow{3}{*}{$CH_{out} \times \frac{R}{2}$ } \\
                 & Normalization [BatchNorm / None] & \\
                 & Non-linearity [ReLU / LeakyReLU] & \\
    \addlinespace
    Refinement 1 & Conv $3$ stride $1$ & \multirow{3}{*}{$CH_{out} \times \frac{R}{2}$ } \\
    {~~~~[OPT]}  & Normalization [BatchNorm / None] & \\
                 & Non-linearity [ReLU / LeakyReLU] & \\
    \addlinespace
    & \hspace{2cm} $\vdots$ & \\
    \addlinespace
    Refinement N & Conv $3$ stride $1$ & \multirow{3}{*}{$CH_{out} \times \frac{R}{2}$ } \\
    {~~~~[OPT]}  & Normalization [BatchNorm / None] & \\
                 & Non-linearity [ReLU / LeakyReLU] & \\
    \midrule
             Params Down. & \multicolumn{2}{c}{$4 \times CH_{in} \times CH_{out} + norm \times 2 \times CH_{out}$} \\
             Params Ref. & \multicolumn{2}{c}{$3 \times CH_{out} \times CH_{out} + norm \times 2 \times CH_{out}$} \\
             Params Tot. & \multicolumn{2}{c}{ $ParamsDown + N \times ParamsRef$} \\
    \bottomrule
  \end{tabular}
  }
\end{center}
\end{table}

\subsection{Architectural composition}
\label{sub:architectural-composition}

The number of building blocks composing the final architecture is determined by the input size $n_{in}$ and the smallest size $n_{out}$ on which the final projection through a linear layer will be performed. Since each building block halves the size of the feature map, the number of blocks used are:

\begin{equation}
    N_b = \left \lfloor log_2 \frac{n_{in}}{n_{out}} \right \rfloor
\end{equation}

where $n_{in}$ and $n_{out}$ must be a power of 2. 

Let $i$ be the position of a generic block in the sequence and let $p_{min}$ be a parameter indicating the output channels of the first block, expressed as a power of two. Let also $p_{max}$ be the maximum number of channels that an activation is allowed to have. Then, each block $b_{i}$ will have a number of filters $F_{b_i}$ defined as:

\begin{equation}
    F_{b_i} = 2^{min(i + p_{min}, p_{max})}
\label{eq:numbers-of-filters-of-the-block}
\end{equation}

The field of view of the last block is then:

\begin{equation}
    FOV = 2N_b\left( 4 + 3N \right)
\end{equation}

Table \ref{tab:ir2v-network-composition} reports the structure of the final convolutional neural network. It includes a pre-processing layer, several building blocks and a projection layer that maps the convolution layers into the output layer.

\begin{table}
\begin{center}
  \caption{Structure of the final convolutional network used in each experiment. $R$ is the spatial resolution of the input signal, $p_{min}$ and $p_{max}$ are two parameters controlling the number of filters of each block and $V$ is the number of output variables.
  }
  \label{tab:ir2v-network-composition}
  \resizebox{.4\textwidth}{!}{ %
  \begin{tabular}{llc}
    \toprule
    Stage & Operation & Output size \\ \midrule
    Pre-processing & Input & $1 \times R$ \\  %
    \addlinespace
    Building blocks & Block 1 & $2^{p_{min}} \times \frac{R}{2}$ \\
             \addlinespace
                    & Block 2 & $2^{1+p_{min}} \times \frac{R}{4}$ \\
             \addlinespace
             \multicolumn{3}{c}{$\vdots$} \\
             \addlinespace
                    & Block N & $2^{N-1+p_{min}} \times \frac{R}{2^N}$ \\
             \addlinespace
             
    Projection & Conv $1 \times 1$ & $V \times 1 \times 1$ \\
               & Flatten & $V$ \\
    \bottomrule
  \end{tabular}
  }
\end{center}
\end{table}

\subsection{Training setup}

The parameters $\theta = \left( \theta_i, \cdots, \theta_{N_b} \right)$ of the system that best explain the training data $\left( x_{train}, y_{train} \right)$ are computed through supervised learning. Let

\begin{equation}
\hat{y} = B^{N_b}(\psi_{N_b}, \theta_{N_b}, B^{N_b-1}(\psi_{N_b-1}, \theta_{N_b-1}, \cdots B^{1}(\psi_1, \theta_1, x_{train})))
\end{equation}

be the predictions of the system over the training set given the input $x_{train}$. The training process aims to find the parameters $\theta$ that minimize the prediction error $min_{\theta} \left(l\left( \hat{y}, y_{train} \right)\right)$ measured through a convex distance loss function $l$.
The selection of the loss function is one of the parameters of the system. We can choose between different functions. The $l1$ loss:

\begin{equation}
    l1(\hat{y}, y_{train}) = \frac{1}{N} \sum_{i=1}^{N} \lvert \hat{y} - y_{train} \rvert
\end{equation}

\noindent as well as the $l2$ loss:

\begin{equation}
    l2(\hat{y}, y_{train}) = \frac{1}{N} \sum_{i=1}^{N} \left( \hat{y} - y_{train} \right)^2
\end{equation}

A third loss combining the advantages of classification and regression is presented in this work. Let $V_i$ be the $i-th$ target variable and $v_i$ one of its values. We reformulate $v_i$ so that it is composed by the sum of two components $c_i$ and $r_i$. $c_i$, here referred as the classification term, is obtained through a non-uniform quantization process where each bin has the same number of values as the splitting is based on quantiles. The regression term $r_i$ is defined as the distance of $v_i$ with respect to the bin, normalized by the distance with the current and the following bin. This decomposition allows us to perform classification on $c_i$ and regression on $r_i$ and, in test time, combine the two quantities to get the estimation of $V_i$.

\section{Dataset} \label{sec:data-composition}
We evaluated the performance of our method on the LUCAS dataset \citep{LUCAS}. It consists in 19,036 topsoil observations from Europe. Each soil sample represents a mixture of soil subsamples taken at each sampling site within 20 cm from the ground surface. For each soil sample, physicochemical and spectral properties are available.

The soil samples composing the LUCAS dataset were air-dried and sieved (2 mm). Then, spectral measurements were collected using an XDS Rapid Content Analyzer (FOSS NIRSystems Inc, Laurel, MD, USA) spectroscope, from 400 to 2500 nm with a spectral sampling interval of 0.5 nm. In addition to spectral measurements, the LUCAS dataset includes 12 chemical and physical soil properties: the percentage of coarse fragments, particle size distribution (\% clay, silt and sand content), pH (in CaCl2 and H2O), organic carbon (g/kg), carbonate content (g/kg), phosphorous content (mg/kg), total nitrogen content (g/kg), extractable potassium content (mg/kg) and the cation exchange capacity (cmol(+)/kg). 

We splitted the library in three sets: train, validation and test so that they contain respectively the $80\%$, the $10\%$ and the $10\%$ of the total amount of samples present in LUCAS dataset. This results in having respectively $13939$, $2000$ and $2000$ samples. The division into subsets was done so that each subset reflects the distribution of the original dataset. 

Table \ref{tab:quantiles} shows that for each variable the quantiles of train, validation and test are the same.

From here on, these sets will be referenced by the notation $\left(x^{set}, y^{set} \right)$ where $set \in \{ train, val, test \}$, $x^{set}$ are the inputs and $y^{set}$ are the expected outputs.

\setlength{\tabcolsep}{4pt}
\begin{table}
\center
\begin{adjustbox}{width=.4\textwidth}
\begin{tabular}{llrrrrrrrrrr}
\toprule
 & set & Q1 & Q2 & Q3 & Q4 & Q5 & Q6 & Q7 & Q8 & Q9 & Q10 \\
\midrule
\parbox[t]{1mm}{\multirow{3}{*}{\rotatebox[origin=c]{90}{coarse}}}
       & tr. & 12\% & 13\% &  6\% & 11\% &  9\% & 11\% &  9\% & 11\% &  9\% &  9\% \\
       & val   & 12\% & 12\% &  6\% & 10\% &  9\% & 12\% & 10\% & 11\% &  9\% &  9\% \\
       & test  & 12\% & 13\% &  6\% & 11\% &  9\% & 11\% & 10\% & 11\% &  8\% &  8\% \\
\midrule
\parbox[t]{1mm}{\multirow{3}{*}{\rotatebox[origin=c]{90}{clay}}}
     & tr. & 13\% &  8\% & 11\% & 11\% &  9\% &  9\% & 10\% & 10\% & 10\% & 10\% \\
     & val   & 13\% &  7\% & 12\% & 11\% &  9\% &  9\% & 11\% & 10\% & 10\% &  9\% \\
     & test  & 12\% &  9\% & 11\% & 12\% &  9\% &  9\% & 10\% & 10\% & 10\% &  9\% \\
\midrule
\parbox[t]{1mm}{\multirow{3}{*}{\rotatebox[origin=c]{90}{silt}}}
     & tr. & 10\% & 11\% &  9\% & 10\% & 10\% & 10\% & 11\% &  9\% & 10\% &  9\% \\
     & val   & 10\% & 10\% &  9\% & 11\% & 11\% & 10\% & 10\% &  9\% & 10\% & 11\% \\
     & test  & 11\% & 10\% &  8\% &  9\% & 10\% &  9\% & 12\% &  9\% & 10\% & 11\% \\
\midrule
\parbox[t]{1mm}{\multirow{3}{*}{\rotatebox[origin=c]{90}{$pH_{CaCl2}$}}}
             & tr. & 10\% & 10\% & 10\% & 10\% & 10\% & 10\% & 10\% & 10\% & 10\% & 10\% \\
             & val   &  9\% & 10\% &  9\% & 10\% & 11\% & 10\% & 10\% & 10\% & 10\% & 11\% \\
             & test  & 10\% &  9\% & 10\% & 11\% & 10\% & 11\% &  9\% & 11\% &  9\% & 11\% \\
\midrule
\parbox[t]{1mm}{\multirow{3}{*}{\rotatebox[origin=c]{90}{$pH_{H2O}$}}}
           & tr. & 10\% & 10\% & 10\% & 10\% & 10\% & 10\% & 10\% & 10\% & 10\% & 10\% \\
           & val   &  9\% &  9\% & 11\% &  9\% & 11\% & 11\% &  9\% & 11\% & 10\% & 11\% \\
           & test  & 11\% &  8\% & 10\% & 11\% & 10\% & 10\% & 10\% & 11\% &  9\% & 11\% \\
\midrule
\parbox[t]{1mm}{\multirow{3}{*}{\rotatebox[origin=c]{90}{OC}}}
   & tr. & 10\% & 10\% & 10\% & 10\% & 10\% & 10\% & 10\% & 10\% & 10\% & 10\% \\
   & val   & 10\% & 11\% &  9\% & 11\% & 10\% &  9\% &  9\% & 10\% & 10\% & 10\% \\
   & test  & 10\% &  9\% & 11\% & 11\% & 11\% & 11\% &  8\% &  9\% &  9\% & 10\% \\
\midrule
\parbox[t]{1mm}{\multirow{3}{*}{\rotatebox[origin=c]{90}{CaCO3}}}
      & tr. & 57\% &  7\% &  6\% & 10\% & 10\% & 10\% \\
      & val   & 32\% & 12\% &  8\% & 14\% & 17\% & 17\% \\
      & test  & 30\% & 12\% &  9\% & 15\% & 17\% & 17\% \\
\midrule
\parbox[t]{1mm}{\multirow{3}{*}{\rotatebox[origin=c]{90}{N}}}
  & tr. & 11\% & 14\% & 11\% & 10\% &  8\% & 10\% &  8\% & 10\% & 10\% & 10\% \\
  & val   & 11\% & 14\% & 11\% &  9\% &  8\% & 10\% &  7\% &  9\% &  9\% & 11\% \\
  & test  & 11\% & 14\% & 10\% & 10\% & 10\% & 11\% &  8\% &  9\% &  9\% &  9\% \\
\midrule
\parbox[t]{1mm}{\multirow{3}{*}{\rotatebox[origin=c]{90}{P}}}
  & tr. & 30\% & 10\% & 10\% & 10\% & 10\% & 10\% & 10\% & 10\% \\
  & val   & 10\% & 13\% & 14\% & 13\% & 12\% & 13\% & 11\% & 15\% \\
  & test  &  9\% & 12\% & 13\% & 13\% & 13\% & 14\% & 14\% & 11\% \\
\midrule
\parbox[t]{1mm}{\multirow{3}{*}{\rotatebox[origin=c]{90}{K}}}
  & tr. & 10\% & 10\% & 10\% & 10\% & 10\% & 10\% & 10\% & 10\% & 10\% & 10\% \\
  & val   &  8\% & 10\% & 10\% & 11\% & 10\% & 11\% & 10\% &  9\% &  9\% & 12\% \\
  & test  & 10\% & 10\% & 10\% &  9\% & 10\% & 10\% & 12\% & 10\% & 10\% & 10\% \\
\midrule
\parbox[t]{1mm}{\multirow{3}{*}{\rotatebox[origin=c]{90}{CEC}}}
    & tr. & 10\% & 10\% & 10\% & 10\% & 10\% & 10\% & 10\% & 10\% & 10\% & 10\% \\
    & val   &  8\% & 10\% &  9\% & 12\% &  9\% & 11\% & 10\% & 11\% & 10\% & 11\% \\
    & test  &  8\% & 11\% &  8\% & 13\% & 10\% & 11\% &  9\% & 11\% & 10\% & 10\% \\
\bottomrule
\end{tabular}

\end{adjustbox}
\caption{Quantiles distribution in training, validation and test.}
\label{tab:quantiles}
\end{table}

\newcommand{\mapgtpic}[1]{\includegraphics[width=.08\linewidth,align=c]{img/maps/lucas/#1.png}}
\newcommand{\mapgtsmallbar}[1]{\includegraphics[width=.08\linewidth,align=c]{img/maps/lucas/small_bars/#1.png}}
\newcommand{\mapgtrow}[1]{ 
\rotatebox[origin=c]{90}{#1}
& \mapgtpic{#1/coarse} & \mapgtpic{#1/clay} & \mapgtpic{#1/silt} & \mapgtpic{#1/sand} & \mapgtpic{#1/pH.in.CaCl2} & \mapgtpic{#1/pH.in.H2O} & \mapgtpic{#1/OC} & \mapgtpic{#1/CaCO3} & \mapgtpic{#1/N} & \mapgtpic{#1/P} & \mapgtpic{#1/K} & \mapgtpic{#1/CEC} \\ }

{

	\setlength{\tabcolsep}{1pt}
	\renewcommand{\arraystretch}{0}

	\begin{figure*}
	\center
	\begin{adjustbox}{width=\textwidth}
	\begin{tabular}{lcccccccccccc}
	& coarse & clay & silt & sand & $pH_{CaCl2}$ &  $pH_{H2O}$ & OC & CaCO3 & N & P & K & CEC \\
	\mapgtrow{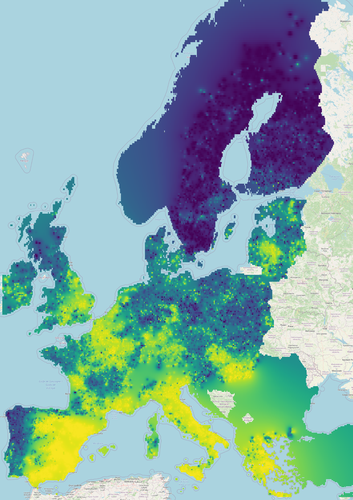} \\
	\mapgtrow{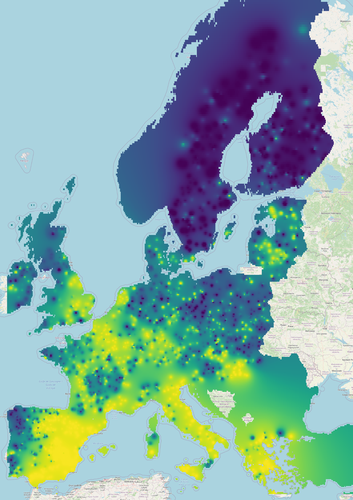} \\
	\mapgtrow{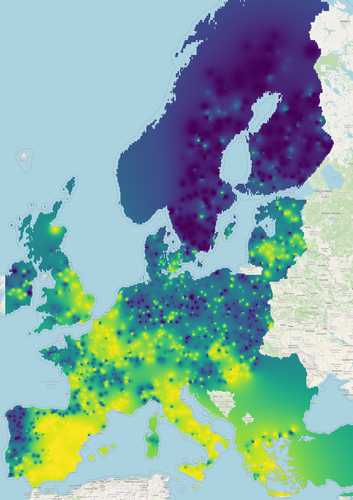} \\

	& \mapgtsmallbar{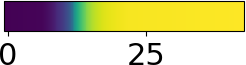} & \mapgtsmallbar{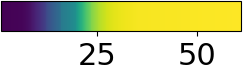} & \mapgtsmallbar{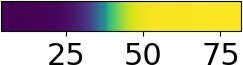} & \mapgtsmallbar{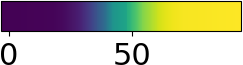} & \mapgtsmallbar{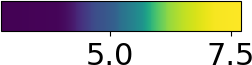} & \mapgtsmallbar{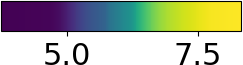} & \mapgtsmallbar{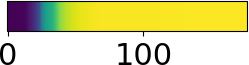} & \mapgtsmallbar{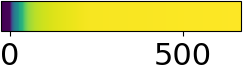} & \mapgtsmallbar{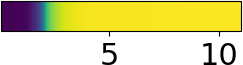} & \mapgtsmallbar{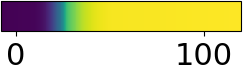} & \mapgtsmallbar{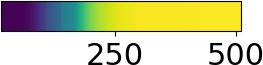} & \mapgtsmallbar{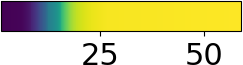} \\

	\end{tabular}
	\end{adjustbox}
	\caption{Train, validation and test maps of each variable composing the LUCAS dataset. These maps have been obtained through IDW interpolation. As it is possible to observe, for each variable the maps are similar, indicating a correct distribution of the samples among train, validation and test set.}
	\label{fig:gt-images}
	\end{figure*}
}

\section{Experiments and Analysis} \label{sec:experiments}

The proposed method has been assessed through several experiments covering all the aspects related to this task, from pure architecture search to model understanding, passing through data analysis. First of all in section~\ref{sub:evaluation_metrics} we discuss the metrics adopted. Then, in section~\ref{sub:sensitivity-analysis} we perform architecture search through the proposed method with the aim of investigating the impact of the different parameters on the regression performance and to define the best configuration for the LUCAS dataset. In subsection \ref{sub:best-result} we present the best result and compare it with the state of the art. We also make a comparison of the best model in predicting all the variables together or each variable at once.
In subsection \ref{sub:feature-importance} we analyze in depth which portion of the spectrum is used in prediction by the best model. 
Finally, in subsection \ref{sub:real-case} we adapt the proposed method to a real case scenario.

\subsection{Evaluation metrics}
\label{sub:evaluation_metrics}

For performance evaluation, we have considered five different metrics. Let $x$ be the output of the system and $y$ the expected output given by the dataset. Then, the mean absolute error is defined as:

\begin{equation}
    MAE(x, y) = \frac{1}{N} \sum_{i=1}{N} \lvert x - y \rvert
\end{equation}

\noindent the mean squared error:

\begin{equation}
    MSE(x, y) = \frac{1}{N} \sum_{i=1}{N} \left( x - y \right)^2
\end{equation}

\noindent the root mean squared error:

\begin{equation}
    RMSE(x, y) = \sqrt{\frac{1}{N} \sum_{i=1}{N} \left( x - y \right)^2}
\end{equation}

\noindent the R2:

\begin{equation}
    R2(x, y) = \frac{\sum_i (y_i-x_i)^2}{\sum_i(y_i-\bar{x})^2}
\end{equation}

\noindent where $\bar{x}$ is the mean value of $x_{i}$ $\forall i$.

The Pearson correlation is defined as follows:

\begin{equation}
  Pearson(x, y) = 
  \frac{ \sum_{i=1}^{n}(x_i-\bar{x})(y_i-\bar{y}) }{%
        \sqrt{\sum_{i=1}^{n}(x_i-\bar{x})^2}\sqrt{\sum_{i=1}^{n}(y_i-\bar{y})^2}}
\end{equation}

\noindent where $\bar{x}$ and $\bar{y}$ are the mean values of $x_{i}$ and  $y_{i}$ $\forall i$ respectively.

\subsection{Sensitivity analysis and architecture search}
\label{sub:sensitivity-analysis}

\begingroup
\newtoggle{inTableHeader}%
\toggletrue{inTableHeader}%
\newcommand*{\StartTableHeader}{\global\toggletrue{inTableHeader}}%
\newcommand*{\EndTableHeader}{\global\togglefalse{inTableHeader}}%

\let\OldTabular\tabular%
\let\OldEndTabular\endtabular%
\renewenvironment{tabular}{\StartTableHeader\OldTabular}{\OldEndTabular\StartTableHeader}%

\newcommand*{\MinNumber}{0.0} %
\newcommand*{\MidNumber}{0.3} %
\newcommand*{\MaxNumber}{1.0} %

\newcommand{\midsepremove}{\aboverulesep = 0mm \belowrulesep = 0mm}
\newcommand{\midsepdefault}{\aboverulesep = 0.605mm \belowrulesep = 0.984mm}

\newcommand{\ApplyGradient}[1]{%
  \iftoggle{inTableHeader}{#1}{
    \ifdim #1 pt > \MidNumber pt
        \pgfmathsetmacro{\PercentColor}{max(min(100.0*(#1 - \MidNumber)/(\MaxNumber-\MidNumber),100.0),0.00)} %
        \hspace{-0.33em}\colorbox{green!\PercentColor!yellow}{\makebox(32,10){#1}}
    \else
        \pgfmathsetmacro{\PercentColor}{max(min(100.0*(\MidNumber - #1)/(\MidNumber-\MinNumber),100.0),0.00)} %
        \hspace{-0.33em}\colorbox{white!\PercentColor!yellow}{\makebox(32,10){#1}}
    \fi
  }}

\newcolumntype{R}{>{\collectcell\ApplyGradient}c<{\endcollectcell}}
\renewcommand{\arraystretch}{0}
\setlength{\fboxsep}{2mm} %
\setlength{\tabcolsep}{0pt}

    \begin{table*}
        \begin{center}
            \begin{adjustbox}{width=\textwidth}
            \midsepremove
            \begin{tabular}{lcRRRRRRRRRRRRR}
              \toprule
               param. & value &  coarse &   clay &    silt &   sand &  $pH_{CaCl2}$ &  $pH_{H2O}$ &     OC &  CaCO3 &      N &       P &      K &    CEC &  global \EndTableHeader\\
               \toprule
               \multirow{3}{*}{$f_{min}$} &       450 & -0.0837 & 0.1991 &  0.0255 & 0.1058 &       0.2657 &     0.2901 & 0.2094 & 0.2111 & 0.1913 & -0.1104 & 0.0029 & 0.1732 &  0.1233 \\
                    &       800 & -0.1025 & 0.1813 &  0.0291 & 0.0851 &       0.2366 &     0.2633 & 0.2231 & 0.2121 & 0.1954 & -0.0785 & 0.0026 & 0.2069 &  0.1212 \\
                    &      1200 & -0.0996 & 0.1930 & -0.0156 & 0.0673 &       0.2662 &     0.2546 & 0.2057 & 0.2115 & 0.1843 & -0.0285 & 0.0032 & 0.1923 &  0.1195 \\
              \midrule
              \multirow{3}{*}{$f_{max}$} &      2500 & -0.0819 & 0.2061 &  0.0417 & 0.0883 &       0.2640 &     0.2550 & 0.2247 & 0.2293 & 0.2130 & -0.0851 &  0.0123 & 0.2194 &  0.1322 \\
                    &      2300 & -0.1151 & 0.2003 &  0.0036 & 0.0582 &       0.2633 &     0.2987 & 0.1989 & 0.1952 & 0.1828 & -0.0356 &  0.0114 & 0.1559 &  0.1182 \\
                    &      2400 & -0.0887 & 0.1671 & -0.0062 & 0.1115 &       0.2411 &     0.2540 & 0.2147 & 0.2104 & 0.1754 & -0.0966 & -0.0149 & 0.1974 &  0.1138 \\
              \midrule
              \multirow{3}{*}{$f_{insz}$} &       512 & -0.0800 & 0.1888 &  0.0331 & 0.0822 &       0.2713 &     0.2599 & 0.2044 & 0.2191 & 0.1979 & -0.0227 & 0.0040 & 0.1813 &  0.1283 \\
                    &      1024 & -0.0881 & 0.1888 & -0.0037 & 0.0946 &       0.2618 &     0.2863 & 0.2176 & 0.2144 & 0.1809 & -0.1318 & 0.0012 & 0.2018 &  0.1187 \\
                    &      2048 & -0.1180 & 0.1958 &  0.0094 & 0.0812 &       0.2350 &     0.2615 & 0.2162 & 0.2010 & 0.1923 & -0.0626 & 0.0034 & 0.1894 &  0.1171 \\
              \midrule
              \multirow{2}{*}{leak} &       0.0 & -0.0748 & 0.1807 &  0.0453 & 0.1003 &       0.2324 &     0.2590 & 0.1923 & 0.1889 & 0.1657 &  0.0059 &  0.0237 & 0.1706 &  0.1242 \\
                    &       0.2 & -0.1159 & 0.2016 & -0.0196 & 0.0717 &       0.2800 &     0.2797 & 0.2332 & 0.2344 & 0.2151 & -0.1512 & -0.0180 & 0.2112 &  0.1185 \\
              \midrule
              \multirow{3}{*}{loss} &             l1 &  0.0244 &  0.3464 &  0.2663 &  0.3039 &       0.4279 &     0.4250 &  0.3217 & 0.3499 &  0.3428 &  0.0819 &  0.0964 &  0.3272 &  0.2761 \\
                    &             l2 &  0.0930 &  0.3181 &  0.1853 &  0.2589 &       0.4373 &     0.4350 &  0.3509 & 0.2369 &  0.3682 &  0.0481 &  0.1151 &  0.3370 &  0.2653 \\
                    & class. & -0.4015 & -0.0900 & -0.4112 & -0.3031 &      -0.0951 &    -0.0507 & -0.0331 & 0.0481 & -0.1383 & -0.3460 & -0.2018 & -0.0903 & -0.1761 \\
              \midrule
              \multirow{2}{*}{b.n.} &      True & -0.0084 &  0.5383 &  0.1894 &  0.3649 &       0.7351 &     0.7393 &  0.5429 &  0.4795 &  0.5124 & -0.0876 &  0.0766 &  0.5173 &  0.3833 \\
                    &     False & -0.1827 & -0.1582 & -0.1647 & -0.1946 &      -0.2258 &    -0.2037 & -0.1195 & -0.0580 & -0.1337 & -0.0570 & -0.0714 & -0.1377 & -0.1422 \\
                   
              \bottomrule
              \end{tabular}
              \end{adjustbox}
        \caption{Sensitivity analysis performed with R2 score. White cells stand for very low R2 values, yellowish cells stand for medium R2 scores, while greenish colors stand for very high R2 values.}
        \label{fig:san_r2}
        \end{center}
    \end{table*}
    \endgroup

The proposed method is highly configurable and thus can be tuned to match the required complexity. In this context, we performed an extensive grid search on LUCAS dataset. 

This search allowed us to discover the best performing architecture on the dataset under analysis. In addition, it permitted us to perform a sensitivity analysis of the system with respect to each parameter. The investigated parameter space is the following:

\begin{itemize} 
\item $f_{min}$ 450, 800, 1200
\item $f_{max}$ 2300, 2400, 2500
\item $f_{insz}$ 512, 1024, 2048
\item \textbf{non-linearity} ReLU, LeakyReLU
\item \textbf{use batchnorm} True, False
\item \textbf{lr} 0.001, 0.0001
\item \textbf{loss} l1,l2, classification
\end{itemize}

where $f_{max}$ and $f_{max}$ represent respectively the starting and ending band under consideration, $f_{insz}$ represents the input size to which the input signal is resampled. The ability to change the spectral boundaries and the signal resolution helps to investigate how important are the spectral bands at the boundaries and how the resolution of the signal affects the overall performance.

Each spectral segnal has been normalized independently so that it has zero mean and unit variance. The ground truth variables of each set are standardized with mean and variances gathered on the training set.

This research space generated 324 setups. Each setup required a training. We have been done 5000 epochs for each training. It involved the use of three nodes connected together. Two machines were equipped with NVIDIA TITAN Xp 12GB while the third one with an NVIDIA GTX 1080 8GB.

The sensivity analysis, with respect to the R2 metric, is summarized in Table \ref{fig:san_r2}. For the sake of clarity, the sensitivity analysis performed with the other metrics has been moved in appendix. Please refer respectively to the Tables \ref{fig:san_mae}, \ref{fig:san_mse}, \ref{fig:san_rmse}, \ref{fig:san_pearson} to view the same analysis performed on MAE, MSE, RMSE, Pearson. For conciseness, these tables do not report all the possible combinations of the network parameters but the behavior of the proposed method with respect to a given parameter at a time.
The table has been color-coded to improve the readability.
White cells stand for very low R2 values, yellowish cells stand for medium R2 scores, while greenish colors indicate very high R2 values. 
This investigation highlighted that the experiments using the full spectrum (i.e. where the first $f_{min}$ and the last considered band $f_{max}$ are respectively 450 and 2500) scored the highest global R2, indicating that all the bands of the spectrum carry information. 
The use of a downsampled version of the spectral signature at a resolution of 512 or 1024 improves the performance (see parameter $f_{insz}$).
This effect is due the lower complexity of the network in presence of a smaller input (see equation \ref{eq:numbers-of-filters-of-the-block}).
The use of ReLU is to be preferred in contrast to the use of LeakyReLU with a leak of 0.2. The loss which scored the best performance is the l1 loss. Finally, the use of batch normalization significatly improves the performances. It is interesting to note that the grid search performed with the other metrics leads to the same results for the use of LeakyReLU as non-linearity, BatchNormalization as a normalization layer and they all tend to use the same the full spectrum information but prefer different input size. MAE and RMSE prefers to have a downsampled version of the spectrum at 512 values, MSE at 1024 while Pearson at 2048. Furthermore, MSE, Pearson and RMSE prefer l2 as a guidance loss in contrast to l1.

\begin{figure*}
\includegraphics[width=\textwidth]{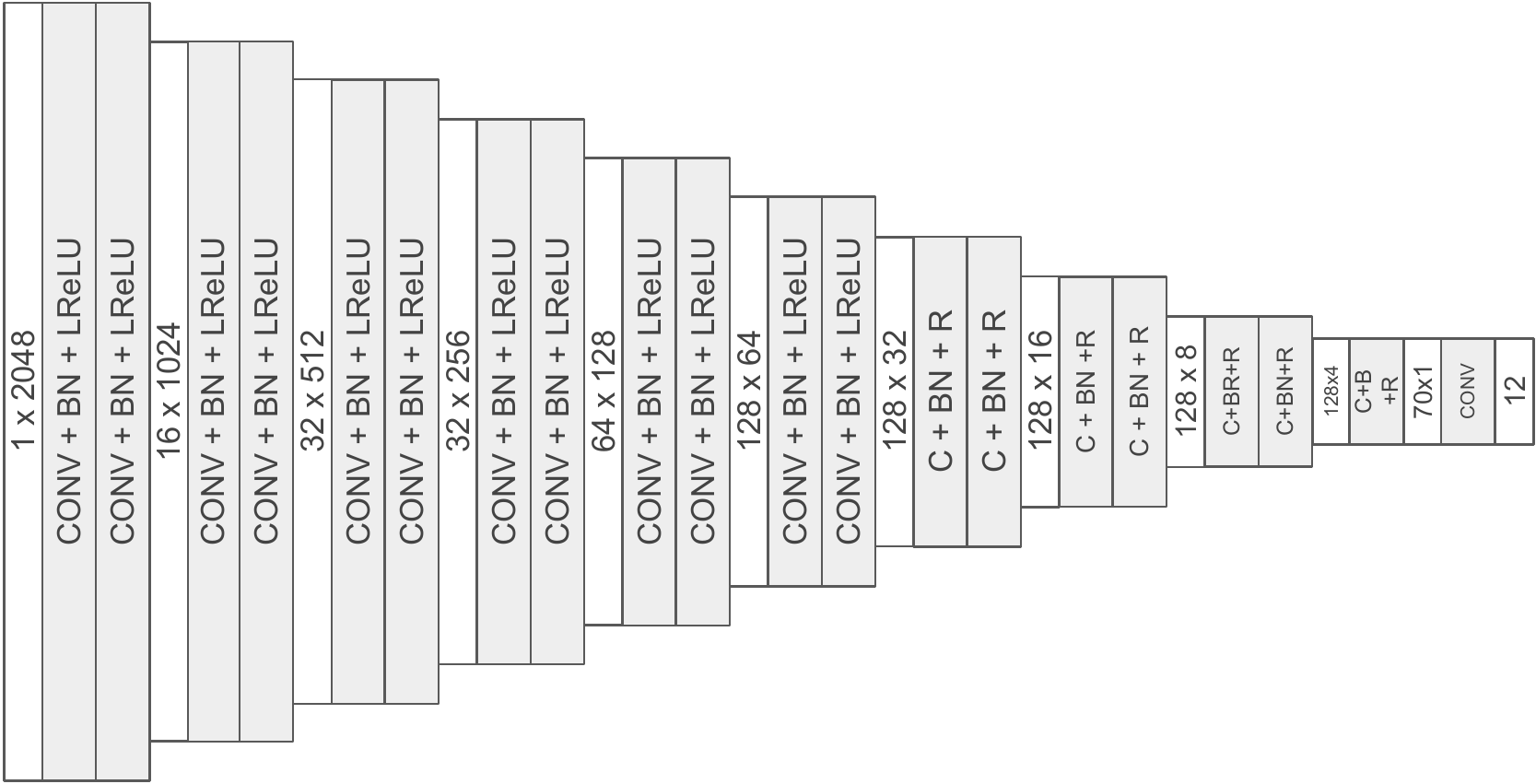}
\caption{Schema of the multi-variable architecture that scored the best R2 in the architecture search process. CONV, as well as C, is a convolutional layer, BN stands for batch normalization and LReLU (or R) stands for Leaky Rectified Linear Unit.}
\label{fig:best-mv-architecture}
\end{figure*}

\subsection{Best result and comparison with the state of the art}
\label{sub:best-result}

The best result in terms of R2 has been achieved using almost the full spectrum (450-2400) resampled at a dimension of 2048. The neural architecture that scored the best result is depicted in figure \ref{fig:best-mv-architecture}. It is composed of by 10 building blocks. Each block has two convolutional layers (CONV or C as in the figure), one for downsampling and one for refinement. Each convolution is followed by a batch normalization layer (BN), with a batch momentum of 0.01, and a leaky rectified linear unit (LReLU or R) with a leak of 0.2. 
The number of channels of the first activation is $2^4$ and maximum number of activations allowed is $2^7$.
The system has been trained with a learning rate of $1e^{-4}$, a weight decay of $0.01$ and the training have been guided by the l1 loss. Table \ref{tab:best-model-mv} shows the best model used in the experiments.

\begin{table}
\begin{center}
  \caption{Architecture of the best model used for multi-variable and single-variable experiments.}
  \label{tab:best-model-mv}
    \resizebox{0.5\textwidth}{!}{ %
  \begin{tabular}{llc}
    \toprule
    Stage & Operation & Output size \\ \midrule
    Pre-processing & Input & $1 \times 2048$ \\
    \addlinespace
    Encoding & Block 1 & $16 \times 1024$ \\
             & Block 2 & $32 \times 512$ \\
             & Block 3 & $64 \times 256$ \\
             & Block 4 & $128 \times 128$ \\
             & Block 5 & $128 \times 64$ \\
             & Block 6 & $128 \times 32$ \\
             & Block 7 & $128 \times 16$ \\
             & Block 8 & $128 \times 8$ \\
             & Block 9 & $128 \times 4$ \\
    \addlinespace
    Projection & Linear+BN+LReLU & $70$ \\
             & Linear & $V$ \\
    \midrule
    Parameters & multi-var ($V=12$) & $723'974$ \\
               & single-var ($V=1$) & $720'097$ \\
    \bottomrule
  \end{tabular}
  }
\end{center}
\end{table}

The performance achieved by this model is shown in Table \ref{fig:variable-scoring-best-solution}. The Table also gives a bright visualization of which variables are more predictable by the best solution, depending by the metric considered. Taking into consideration the R2 metric (fourth column), it's possible to observe how the PH in $CaCl2$ can be predicted from the spectral signature with a very good reliability (0.9175 of R2). Among all the variables considered in this work, only the coarse and the potassium can be predicted with an R2 lower than 0.3. 

\begin{figure*}
\includegraphics[width=\textwidth]{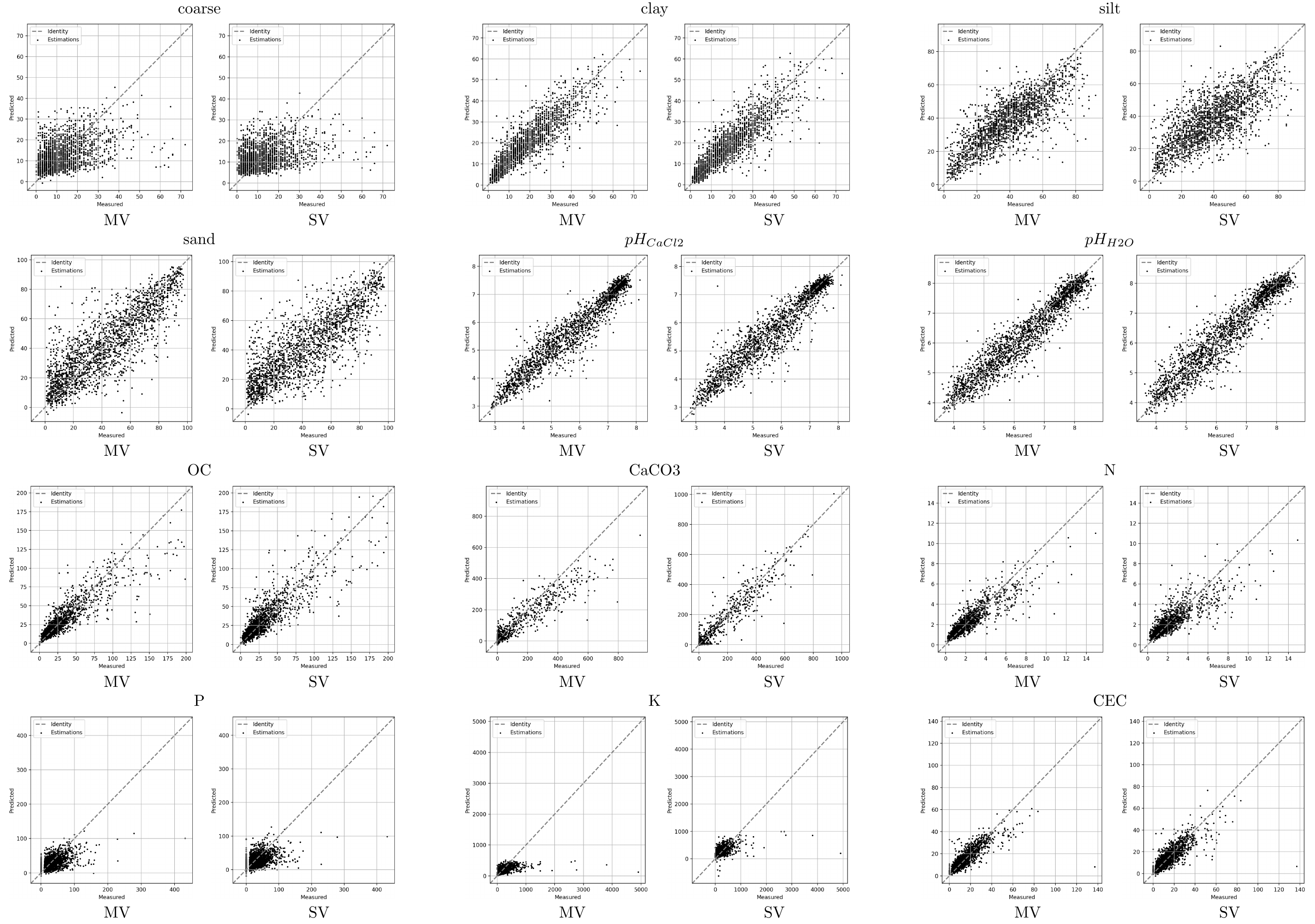}
\caption{Scatter plots of ground-truths vs predicted values.}
\label{fig:scatter-plots}
\end{figure*}

Figure \ref{fig:scatter-plots} shows, for each considered variable, the scatter plots of the estimations obtained by the presented method, compared with the ground truths. The relationship among the spectral signals and the variables is surjective. Specifically, different hyperspectral signals correspond to a specific value of a variable.
In table \ref{tab:average-number-of-spectra-for-each-value} is in fact possible to observe the average number of spectra associated to each unique value for each variable. Interestingly, this number affects the prediction performance for a variable as the task becomes harder.

These results demonstrate that it's possible to infer these variables from the spectral signature. Some readers can object that in this context the estimation of a variable can be brought by the codependency with another target variable and therefore it's not said that there is a relationship with the hyperspectral signature. To make sure that this does not occur in this case, we also performed other experiments with the configuration of the best model, but instead of predicting all the variable at once, we predict at each time a different variable, so that for example the model in charge of predicting the organic carbon is unaware of the values of the other variables. Table \ref{tab:comparison-with-sota} reports the number of these experiments. The row ``ours [mv]'' reports the results of the best multivariable estimation method while ``ours [sv]'' reports the results achieved by our method trained to estimate each variable separately. As expected, the performances of the models trained separately have a lower score, however the difference with the multi variable model is small (0.05 of difference in terms of R2), suggesting that all variables can be directly inferred by the hyperspectral signal.

\begin{table*}
     \centering
     \begin{subtable}[t]{0.17\textwidth}
         \centering
            \begin{tabular}{lr}
                \toprule
                var & value \\
                \midrule
                clay & 0.0189 \\
                coarse & 0.0263 \\
                $pH_{CaCl2}$ & 0.0324 \\
                OC & 0.0327 \\
                N & 0.0408 \\
                $pH_{H2O}$ & 0.0427 \\
                K & 0.0447 \\
                silt & 0.0751 \\
                P & 0.0800 \\
                CEC & 0.0830 \\
                sand & 0.0962 \\
                CaCO3 & 0.0962 \\
                \midrule
                \textbf{avg} & \textbf{0.0567} \\
            \end{tabular}
            \caption{MAE}
     \end{subtable}
     \hfill
     \begin{subtable}[t]{0.17\textwidth}
         \centering
            \begin{tabular}{lr}
                \toprule
                var & value \\
                \midrule
                clay & 0.0011 \\
                coarse & 0.0014 \\
                $pH_{CaCl2}$ & 0.0025 \\
                OC & 0.0031 \\
                N & 0.0053 \\
                $pH_{H2O}$ & 0.0078 \\
                K & 0.0079 \\
                silt & 0.0186 \\
                P & 0.0192 \\
                CEC & 0.0195 \\
                sand & 0.0272 \\
                CaCO3 & 0.0320 \\
                \midrule
                \textbf{avg} & \textbf{0.0158} \\
            \end{tabular}
            \caption{MSE}
     \end{subtable}
     \hfill
     \begin{subtable}[t]{0.17\textwidth}
         \centering
            \begin{tabular}{lr}
                \toprule
                var & value \\
                \midrule
                $pH_{CaCl2}$ & 0.9583 \\
                $pH_{H2O}$ & 0.9548 \\
                CaCO3 & 0.9456 \\
                OC & 0.8924 \\
                N & 0.8838 \\
                clay & 0.8827 \\
                CEC & 0.8509 \\
                sand & 0.8370 \\
                silt & 0.7898 \\
                P & 0.5885 \\
                K & 0.4865 \\
                coarse & 0.4491 \\
                \midrule
                \textbf{avg} & \textbf{0.7933} \\
            \end{tabular}
            \caption{Pearson}
     \end{subtable}
     \hfill
     \begin{subtable}[t]{0.17\textwidth}
         \centering
            \begin{tabular}{lr}
                \toprule
                var & value \\
                \midrule
                $pH_{CaCl2}$ & 0.9175 \\
                $pH_{H2O}$ & 0.9094 \\
                CaCO3 & 0.8544 \\
                OC & 0.7898 \\
                N & 0.7741 \\
                clay & 0.7701 \\
                CEC & 0.7222 \\
                sand & 0.6985 \\
                silt & 0.6199 \\
                P & 0.3434 \\
                K & 0.2213 \\
                coarse & 0.1487 \\
                \midrule
                \textbf{avg} & \textbf{0.6474} \\
            \end{tabular}
            \caption{R2}
     \end{subtable}
     \hfill
     \begin{subtable}[t]{0.17\textwidth}
         \centering
            \begin{tabular}{lr}
                \toprule
                var & value \\
                \midrule
                clay & 0.0335 \\
                coarse & 0.0373 \\
                $pH_{CaCl2}$ & 0.0503 \\
                OC & 0.0554 \\
                N & 0.0728 \\
                $pH_{H2O}$ & 0.0885 \\
                K & 0.0886 \\
                silt & 0.1363 \\
                P & 0.1386 \\
                CEC & 0.1398 \\
                sand & 0.1649 \\
                CaCO3 & 0.1788 \\
                \midrule
                \textbf{avg} & \textbf{0.1015} \\
            \end{tabular}
            \caption{RMSE}
     \end{subtable}
\caption{Variable scoring according to the different metrics of the best solution.}
\label{fig:variable-scoring-best-solution}
\end{table*}

\begin{table*}
\begin{adjustbox}{width=\textwidth}
\begin{tabular}{lrrrrrrrrrrrrr}
\toprule
 method & coarse &   clay &    silt &   sand &  $pH_{CaCl2}$ &  $pH_{H2O}$ &     OC &  CaCO3 &      N &       P &      K &    CEC &  avg \\
 \midrule
RF & 0.0942 & 0.478 & 0.3961 & 0.3996 & 0.453 & 0.4737 & 0.5304 & 0.5454 & 0.4408 & 0.0902 & 0.1422 & 0.4069 & 0.3709 \\
SVR & \textbf{0.1811} & 0.7402 & 0.5222 & 0.5824 & 0.8494 & 0.8441 & 0.7219 & 0.7591 & 0.7198 & 0.1455 & 0.1386 & 0.6305 & 0.5696 \\
BRT & 0.1279 & 0.5054 & 0.3998 & 0.4132 & 0.5556 & 0.5753 & 0.6001 & 0.6669 & 0.4728 & 0.096 & 0.1133 & 0.4480 & 0.4145 \\
 
 \textbf{ours [mv]} & 0.1487 & \textbf{0.7701} & \textbf{0.6199} & \textbf{0.6985} & \textbf{0.9175} & \textbf{0.9094} & \textbf{0.7898} & \textbf{0.8544} & \textbf{0.7741} & \textbf{0.3434} & \textbf{0.2213} & \textbf{0.7222} & \textbf{0.6474} \\
 ours [sv] & 0.1095 & 0.7626 & 0.5243 & 0.6592 & 0.9032 & 0.8999 & 0.7637 & 0.9213 & 0.6904 & 0.2315 & 0.3433 & 0.3433 & 0.5960 \\
 \bottomrule
\end{tabular}
\end{adjustbox}
\caption{Comparison with the state of the art (R2)}
\label{tab:comparison-with-sota}
\end{table*}

Table  \ref{tab:comparison-with-sota} reports also the comparison with the state of the art that is represented by Random Forests (RF), Support Vector Machines (SVR) and Boosting Regression Trees (BRT). These are the main techniques used in the methods belonging to the state of art, adapted to the current domain. RF are used by \cite{Forkuor, zhou2021prediction, hu2019quantitative, meng}, SVR are used by \cite{Forkuor, zhou2021prediction, meng} and BRT are used by \cite{Forkuor, zhou2021prediction}. For a fair comparison, a grid search on the parameters of each method as been conduced.
Beside numerical assessment, we furnish to the reader a visualization of the results of our method and the methods belonging to the state of the art for a fast and meaningful visual comparison with the ground truth. This is done by interpolating the predictions through the inverse distance weighting (IDW) interpolation. Figure \ref{fig:map-big-pH.in.H2O} shows a visual comparison of the different methods with the ground truth for the $pH$ in $H_20$. Figures for $pH$ in $CaCl2$ (fig. \ref{fig:map-big-pH.in.CaCl2}), $CaCO3$ (fig. \ref{fig:map-big-CaCO3}), $OC$ (fig. \ref{fig:map-big-OC}), $N$ (fig. \ref{fig:map-big-N}), $clay$ (fig. \ref{fig:map-big-clay}), $CEC$ (fig. \ref{fig:map-big-CEC}), $sand$ (fig. \ref{fig:map-big-sand}), $silt$ (fig. \ref{fig:map-big-silt}), $P$ (fig. \ref{fig:map-big-P}), $K$ (fig. \ref{fig:map-big-K}), $coarse$ (fig. \ref{fig:map-big-coarse}) can be found in the appendix. Figure \ref{fig:prediction-maps} shows the predictions of the multi-variable configuration.

\newcolumntype{R}[2]{%
    >{\adjustbox{angle=#1,lap=\width-(#2)}\bgroup}%
    l%
    <{\egroup}%
}
\newcommand*\rot{\multicolumn{1}{R{90}{1em}}}%

\begin{table}
\begin{tabular}{cccccccccccc}
\toprule
\rot{coarse} &  \rot{clay} &  \rot{silt} &  \rot{sand} &  \rot{$pH_{CaCl2}$} &  \rot{$pH_{H2O}$} &   \rot{OC} &  \rot{CaCO3} &     \rot{N} &    \rot{P} &    \rot{K} &  \rot{CEC} \\
\midrule
33 & 32 & 23 & 20 & 4 &  4 & 3 & 6 & 22 & 3 & 1 & 5 \\
\bottomrule
\end{tabular}
\caption{Average number of different spectra associated to a specific value of each variable. Variables with the tendency to be bijective are more easy to estimate.}
\label{tab:average-number-of-spectra-for-each-value}
\end{table}

\newcommand{\mapbigpic}[1]{\includegraphics[width=.50\linewidth]{img/maps/#1.png}}

\newcommand{\mapbig}[2]{
\begin{figure*}
    \setlength{\tabcolsep}{1pt}
    \renewcommand{\arraystretch}{.5}
    \centering
    \begin{tabular}{cc}
        \mapbigpic{gt/#1} & \mapbigpic{mv/#1} \\
        ground truth & multi-variable pred. \\
        \mapbigpic{sv/#1} & \mapbigpic{rf/#1} \\
        single-variable pred. & random forest \\
        \mapbigpic{svr/#1} & \mapbigpic{brt/#1}  \\
        \multicolumn{2}{c}{\includegraphics[width=\linewidth]{img/maps/bars/#1.png}} \\
        svr & brt \\
    \end{tabular}
\caption{Visual comparison of the proposed methods with the state of the art for the variable #2.}
    \label{fig:map-big-#1}
\end{figure*}
}

\mapbig{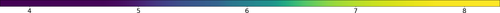}{$pH_{H_2O}$}

\newcommand{\maprespic}[1]{\includegraphics[width=.34\linewidth]{img/maps/mv/#1.png}}
\newcommand{\mapresbar}[1]{\includegraphics[width=.34\linewidth]{img/maps/mv/bars/#1.png}}

\newcommand{\mapresrow}[6]{
    \maprespic{#1} & \maprespic{#3} & \maprespic{#5} \\
    \mapresbar{#1} & \mapresbar{#3} & \mapresbar{#5} \\
    #2 & #4 & #6 \\
}

\begin{figure*}
    \hspace*{-.7cm}
    \setlength{\tabcolsep}{3pt} %
    \renewcommand{\arraystretch}{.5}
    \centering
    \begin{tabular}{cccccc}
        \mapresrow{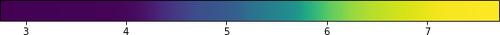}{$pH_{CaCl_2}$}{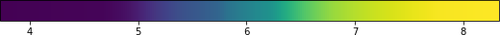}{$pH_{H_2O}$}{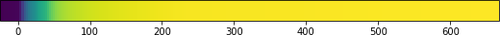}{CaCO3}
        \mapresrow{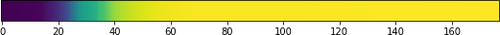}{OC}{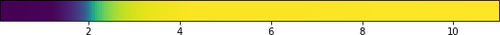}{N}{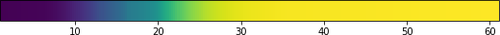}{clay}
        \mapresrow{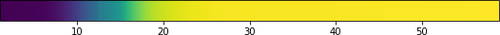}{CEC}{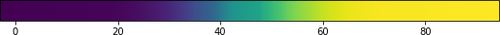}{sand}{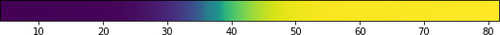}{silt}
        \mapresrow{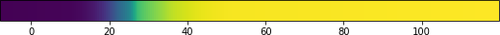}{P}{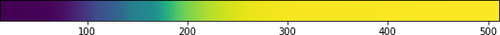}{K}{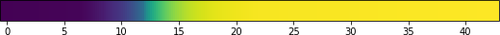}{coarse}
    \end{tabular}
\caption{Predictions of the multi-variable configuration on all the considered variables.}
\label{fig:prediction-maps}
\end{figure*}

\subsection{Real case scenario}
\label{sub:real-case}

\begin{table*}
\begin{adjustbox}{width=\textwidth}

    \begin{tabular}{lrrrrrrrrrrrrr}
    \toprule
    method &  coarse &   clay &   silt &   sand &  $pH_{CaCl2}$ &  $pH_{H2O}$ &     OC &  CaCO3 &      N &      P &      K &    CEC & avg \\
    \toprule
       brt &  0.0976 & 0.4512 & 0.3964 & 0.3923 &       0.4874 &     0.5112 & 0.5595 & 0.5997 & 0.4072 & 0.0560 & 0.0506 & 0.3645 & 0.3645 \\
        rf &  0.0916 & 0.4236 & 0.4012 & 0.3890 &       0.3760 &     0.3942 & 0.5149 & 0.5194 & 0.4178 & 0.0968 & 0.1140 & 0.3346 & 0.3394 \\
       svr &  0.0649 & 0.4709 & 0.3976 & 0.4101 &       0.7215 &     0.7315 & 0.5587 & 0.1034 & 0.5305 & 0.0618 & 0.0536 & 0.4087 & 0.3761 \\
      \textbf{ours [mv]} & 0.1521 & 0.7453 & 0.6305 & 0.6774 &        0.8742 &     0.8648 & 0.7642 & 0.8100 & 0.7191 & 0.2703 & 0.2307 & 0.6599 & \textbf{0.6165} \\
    \bottomrule
    \end{tabular}

\end{adjustbox}
\caption{R2 score obtained in th real case scenario. BRT, RF, SVR are respectively the Boosting Regression Trees, the Random Forests and the Support Vector Machines.}
\label{fig:real-case-scenario}
\end{table*}

To test the performances of the proposed method on PRISMA hyperspectral data in ideal conditions, the LUCAS datset was resampled assuming Gaussian spectral response functions to the specific spectral response of the PRISMA~\citep{loizzo2018prisma} sensor.
Bands from 1338.9 nm to 1501.7 nm and from 1784.4 nm to 1993.2 nm have been removed because located in the atmospheric water absorption bands. This results in an ideal PRISMA spectral dataset consisting in 170 spectral band each.
The best configuration in this case is to use the spectrum between 400 and 2500 resampled at a dimension of 128. A minimum dimension before projection of 4, which generates a set of 5 blocks. The first block having 64 filter banks while the following ones have twice as much as the previous block, while producing a feature map of half of the size. A limit on the maximum number of allowed filters is set to 128. The remaining parameters are the same as the ones found in the previous experiments for multi-variable prediction on laboratory data. Figure \ref{tab:best-model-real-case} shows the found architecture for the real case scenario.
Results are shown in table \ref{fig:real-case-scenario}. As expected the performance drop with the introduced deterioration. Specifically, the proposed method loses a delta of 0.0309 in terms of R2. We believe that our approach can be implemented on real data such as PRISMA. In contrast, methods representing the state of the art have a huge drop. This indicates that the proposed solution is much more robust to degradation.

\begin{table}
\begin{center}
  \caption{Architecture of the best model used for the real case scenario.}
  \label{tab:best-model-real-case}
  \resizebox{0.5\textwidth}{!}{ %
  \begin{tabular}{llc}
    \toprule
    Stage & Operation & Output size \\ \midrule
    Pre-processing & Input & $1 \times 128$ \\
    \addlinespace
    Encoding & Block 1 & $16 \times 64$ \\
             & Block 2 & $32 \times 32$ \\
             & Block 3 & $64 \times 16$ \\
             & Block 4 & $128 \times 8$ \\
             & Block 5 & $128 \times 4$ \\
    \addlinespace
    Projection & Linear+BN+LReLU & $70$ \\
             & Linear & $12$ \\
    \midrule
    Parameters & multi-var & $262'150$ \\
    \bottomrule
  \end{tabular}
  }
\end{center}
\end{table}

\subsection{Feature importance}
\label{sub:feature-importance}
\begin{figure*}
    \centering
    \includegraphics[width=1\linewidth]{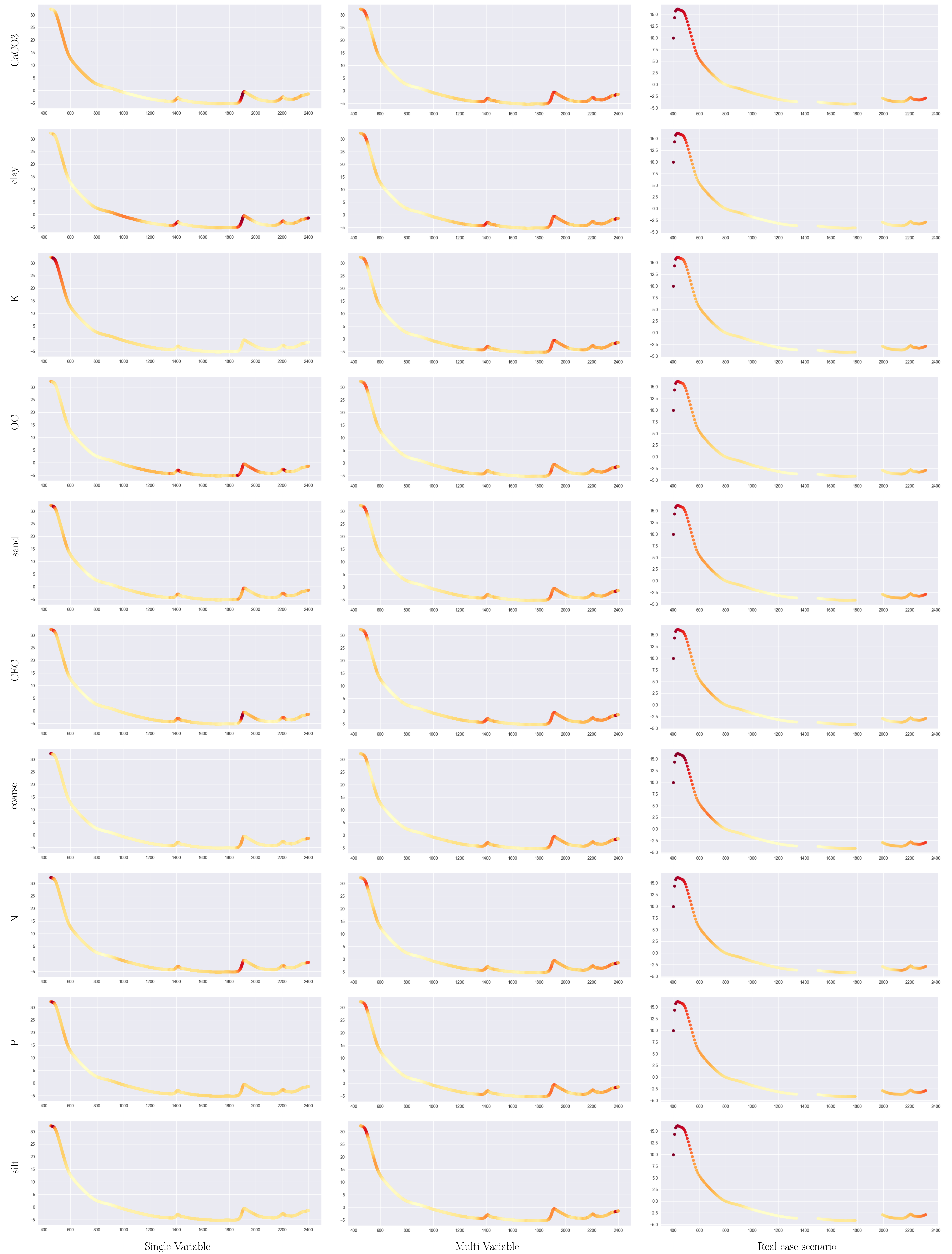}
    \caption{Feature importance. Reddish colors show the frequencies of the spectrum that the proposed network uses for the estimation of the soil variable. Here we report the feature importance for the single and multi variable approach as well as for the real case scenario. }
    \label{fig:feature-importance}
\end{figure*}

In this experiment we analyze in depth the different configurations to assess which parts of the spectral information are more relevant. In particular, we are interested to evaluate the differences - if any - among single- and multi- variable prediction and from data acquired in laboratory versus simulated real samples. 
This investigation has been conduced through the use of GradCAM \citep{gradcam}, averaged on all the samples in the test set. For visualization purposes, in figure \ref{fig:feature-importance} the heatmaps have been drawn on the first spectrum of the testset. As it is possible to observe on single variable and multivariable configurations the first part of the spectra (<600) and the peaks around 1400 and 1900 carry the most important parts of information for predicting the chemicals from spectral information. These findings are compatible with the ones found by \cite{wang2022using} and \cite{vohland2017quantification}. 
Note that the water absorptions can be included in the evaluated bands as  data-driven methods can take in consideration the context of the important bands.
The multi-variable prediction is more conservative and tends to evaluate always these three portions while predicting each output, while the single-variable prediction setup gets more specialized and therefore offers a better overview of which part is really important in estimating each variable independently. In particular, the prediction of CaCO3, clay, sand, CEC, coarse, N heavily depends on all the three defined salient parts, while the prediction of K, P, silt is mostly entrusted in the initial part of the spectra and finally the prediction of the organic carbon depends heavily from the two peaks. And since precisely these two peaks are lost in the degradation of the real case scenario, the model uses the second most important part of the signal, which is its beginning. This of course results in a drop of the performance.

\section{Conclusions} \label{sec:conclusions}

The use of machine and deep learning for estimating soil parameters from remote measurements such as hyperspectral measurements is widespread. Several approaches in the state of the art address the problem but they present some limitations. 

In this paper we proposed a flexible neural architecture for the prediction of multiple soil properties at once that allows a fast and scalable architecture search. This permitted to overcome one of the limitations in the state of the art since most of the methods estimate one variable at time. 

To test and validate our proposed approach we considered the LUCAS dataset ~\citep{LUCAS} that consists in 19,036 topsoil observations from Europe and it includes 12 soil parameters and the corresponding hyperspectral measure. 

We have conduced a grid search to perform sensitivity analysis and architecture search. The outcome of the latter experiment assessed the importance of each parameter of the system, while the latter offered the best setup for the considered dataset.

Afterwards we evaluated the drop in performance in using the best setup to predict each variable at once and we showed that this drop is not relevant. To assess the performance of the system in a real context, we created a synthetic dataset that simulates the spectral signals of Prisma and its degradation. Afterwards we compared the performance of the proposed approach on this data and we showed that the proposed method is more robust to degradation in contrast to the methods belonging to the state of the art. We went deeper and we furnished relevant visual information that help to better understand all the aspects related to this research. In first place, we proposed a visual interpolation of the outputs to make a visual comparison among the proposed solutions and the models of the state of the art. Finally, we investigated the importance of the bands of the spectral signal that the different setups evaluate in predicting each soil characteristic.

As a future work the idea is to work in two directions. From one side we wish to add to our input other measures such as the features extracted from the Digital Elevation Model (\cite{mammadov2021estimation, khanal2018integration}).
On the other side we wish to experiment the powerful of our approach on real aerial hyperspectral data. This last task is not easy since it is very difficult to collect at the same time soil parameters with proximal measures and aerial hyperspectral measures. However, we are working on the collection of this data in the context of the Pignoletto project (more info can be found here \href{https://pignoletto.mib.infn.it/index.php/it/}{https://pignoletto.mib.infn.it/index.php/it/}).

\section{Code availability}
The code of this paper can be found 
\href{https://github.com/dros1986/scalable-cnn-for-soil-properties-estimation}{at the address in footnote}
\footnote[1]{\href{https://github.com/dros1986/scalable-cnn-for-soil-properties-estimation}{https://github.com/dros1986/scalable-cnn-for-soil-properties-estimation}}. It is developed in Python3. Further details on how to use it can be found on the repository.

\section*{Acknowledgement}
This research has been conduced in the context of PIGNOLETTO project $-$ Call HUB Ricerca e Innovazione (CUP number E41B20000050007), co-funded by Regional Operational Programme, European Regional Development Fund (POR FESR 2014-2020).

\printcredits

\bibliographystyle{elsarticle-harv}

\bibliography{cas-refs}

\begin{thebibliography}{33}
\expandafter\ifx\csname natexlab\endcsname\relax\def\natexlab#1{#1}\fi
\providecommand{\url}[1]{\texttt{#1}}
\providecommand{\href}[2]{#2}
\providecommand{\path}[1]{#1}
\providecommand{\DOIprefix}{doi:}
\providecommand{\ArXivprefix}{arXiv:}
\providecommand{\URLprefix}{URL: }
\providecommand{\Pubmedprefix}{pmid:}
\providecommand{\doi}[1]{\href{http://dx.doi.org/#1}{\path{#1}}}
\providecommand{\Pubmed}[1]{\href{pmid:#1}{\path{#1}}}
\providecommand{\bibinfo}[2]{#2}
\ifx\xfnm\relax \def\xfnm[#1]{\unskip,\space#1}\fi
\bibitem[{Angelopoulou et~al.(2019a)Angelopoulou, Tziolas, Balafoutis, Zalidis
  and Bochtis}]{angelopoulou2019remote}
\bibinfo{author}{Angelopoulou, T.}, \bibinfo{author}{Tziolas, N.},
  \bibinfo{author}{Balafoutis, A.}, \bibinfo{author}{Zalidis, G.},
  \bibinfo{author}{Bochtis, D.}, \bibinfo{year}{2019}a.
\newblock \bibinfo{title}{Remote sensing techniques for soil organic carbon
  estimation: A review}.
\newblock \bibinfo{journal}{Remote Sensing} \bibinfo{volume}{11},
  \bibinfo{pages}{676}.
\bibitem[{Angelopoulou et~al.(2019b)Angelopoulou, Tziolas, Balafoutis, Zalidis
  and Bochtis}]{angelopoulou}
\bibinfo{author}{Angelopoulou, T.}, \bibinfo{author}{Tziolas, N.},
  \bibinfo{author}{Balafoutis, A.}, \bibinfo{author}{Zalidis, G.},
  \bibinfo{author}{Bochtis, D.}, \bibinfo{year}{2019}b.
\newblock \bibinfo{title}{Remote sensing techniques for soil organic carbon
  estimation: A review}.
\newblock \bibinfo{journal}{Remote Sensing} \bibinfo{volume}{11},
  \bibinfo{pages}{676}.
\newblock \DOIprefix\doi{10.3390/rs11060676}.
\bibitem[{Ben-Dor(2002)}]{ben2002quantitative}
\bibinfo{author}{Ben-Dor, E.}, \bibinfo{year}{2002}.
\newblock \bibinfo{title}{Quantitative remote sensing of soil properties} .
\bibitem[{Castaldi et~al.(2016)Castaldi, Palombo, Santini, Pascucci, Pignatti
  and Casa}]{Castaldi2016}
\bibinfo{author}{Castaldi, F.}, \bibinfo{author}{Palombo, A.},
  \bibinfo{author}{Santini, F.}, \bibinfo{author}{Pascucci, S.},
  \bibinfo{author}{Pignatti, S.}, \bibinfo{author}{Casa, R.},
  \bibinfo{year}{2016}.
\newblock \bibinfo{title}{Evaluation of the potential of the current and
  forthcoming multispectral and hyperspectral imagers to estimate soil texture
  and organic carbon}.
\newblock \bibinfo{journal}{Remote Sensing of Environment2}
  \bibinfo{volume}{179}, \bibinfo{pages}{54--65}.
\newblock \URLprefix
  \url{https://www.sciencedirect.com/science/article/pii/S0034425716301195},
  \DOIprefix\doi{https://doi.org/10.1016/j.rse.2016.03.025}.
\bibitem[{Cogliati et~al.(2021)Cogliati, Sarti, Chiarantini, Cosi, Lorusso,
  Lopinto, Miglietta, Genesio, Guanter, Damm et~al.}]{cogliati2021prisma}
\bibinfo{author}{Cogliati, S.}, \bibinfo{author}{Sarti, F.},
  \bibinfo{author}{Chiarantini, L.}, \bibinfo{author}{Cosi, M.},
  \bibinfo{author}{Lorusso, R.}, \bibinfo{author}{Lopinto, E.},
  \bibinfo{author}{Miglietta, F.}, \bibinfo{author}{Genesio, L.},
  \bibinfo{author}{Guanter, L.}, \bibinfo{author}{Damm, A.}, et~al.,
  \bibinfo{year}{2021}.
\newblock \bibinfo{title}{The prisma imaging spectroscopy mission: Overview and
  first performance analysis}.
\newblock \bibinfo{journal}{Remote Sensing of Environment}
  \bibinfo{volume}{262}, \bibinfo{pages}{112499}.
\bibitem[{Crucil et~al.(2019)Crucil, Castaldi, Aldana-Jague, van Wesemael,
  Macdonald and Van~Oost}]{crucil2019assessing}
\bibinfo{author}{Crucil, G.}, \bibinfo{author}{Castaldi, F.},
  \bibinfo{author}{Aldana-Jague, E.}, \bibinfo{author}{van Wesemael, B.},
  \bibinfo{author}{Macdonald, A.}, \bibinfo{author}{Van~Oost, K.},
  \bibinfo{year}{2019}.
\newblock \bibinfo{title}{Assessing the performance of uas-compatible
  multispectral and hyperspectral sensors for soil organic carbon prediction}.
\newblock \bibinfo{journal}{Sustainability} \bibinfo{volume}{11},
  \bibinfo{pages}{1889}.
\bibitem[{Dombi and Dineva(2020)}]{dombi2020adaptive}
\bibinfo{author}{Dombi, J.}, \bibinfo{author}{Dineva, A.},
  \bibinfo{year}{2020}.
\newblock \bibinfo{title}{Adaptive savitzky-golay filtering and its
  applications}.
\newblock \bibinfo{journal}{International Journal of Advanced Intelligence
  Paradigms} \bibinfo{volume}{16}, \bibinfo{pages}{145--156}.
\bibitem[{Forkuor et~al.(2017)Forkuor, Hounkpatin, Welp and Thiel}]{Forkuor}
\bibinfo{author}{Forkuor, G.}, \bibinfo{author}{Hounkpatin, O.K.L.},
  \bibinfo{author}{Welp, G.}, \bibinfo{author}{Thiel, M.},
  \bibinfo{year}{2017}.
\newblock \bibinfo{title}{High resolution mapping of soil properties using
  remote sensing variables in south-western burkina faso: A comparison of
  machine learning and multiple linear regression models}.
\newblock \bibinfo{journal}{PLOS ONE} \bibinfo{volume}{12},
  \bibinfo{pages}{1--21}.
\newblock \URLprefix \url{https://doi.org/10.1371/journal.pone.0170478},
  \DOIprefix\doi{10.1371/journal.pone.0170478}.
\bibitem[{Galeazzi et~al.(2008)Galeazzi, Sacchetti, Cisbani and
  Babini}]{galeazzi2008prisma}
\bibinfo{author}{Galeazzi, C.}, \bibinfo{author}{Sacchetti, A.},
  \bibinfo{author}{Cisbani, A.}, \bibinfo{author}{Babini, G.},
  \bibinfo{year}{2008}.
\newblock \bibinfo{title}{The prisma program}, in: \bibinfo{booktitle}{IGARSS
  2008-2008 IEEE International Geoscience and Remote Sensing Symposium},
  \bibinfo{organization}{IEEE}. pp. \bibinfo{pages}{IV--105}.
\bibitem[{Gebbers and Adamchuk(2010)}]{gebbers2010precision}
\bibinfo{author}{Gebbers, R.}, \bibinfo{author}{Adamchuk, V.I.},
  \bibinfo{year}{2010}.
\newblock \bibinfo{title}{Precision agriculture and food security}.
\newblock \bibinfo{journal}{Science} \bibinfo{volume}{327},
  \bibinfo{pages}{828--831}.
\bibitem[{Guo et~al.(2019)Guo, Zhang, Shi, Chen, Jiang and
  Linderman}]{guo2019prediction}
\bibinfo{author}{Guo, L.}, \bibinfo{author}{Zhang, H.}, \bibinfo{author}{Shi,
  T.}, \bibinfo{author}{Chen, Y.}, \bibinfo{author}{Jiang, Q.},
  \bibinfo{author}{Linderman, M.}, \bibinfo{year}{2019}.
\newblock \bibinfo{title}{Prediction of soil organic carbon stock by laboratory
  spectral data and airborne hyperspectral images}.
\newblock \bibinfo{journal}{Geoderma} \bibinfo{volume}{337},
  \bibinfo{pages}{32--41}.
\bibitem[{Heil and Schmidhalter(2021)}]{heil2021evaluation}
\bibinfo{author}{Heil, K.}, \bibinfo{author}{Schmidhalter, U.},
  \bibinfo{year}{2021}.
\newblock \bibinfo{title}{An evaluation of different nir-spectral
  pre-treatments to derive the soil parameters c and n of a humus-clay-rich
  soil}.
\newblock \bibinfo{journal}{Sensors} \bibinfo{volume}{21},
  \bibinfo{pages}{1423}.
\bibitem[{Hofer et~al.(2006)Hofer, Kaufmann, Stuffler, Penn{\'e}, Schreier,
  M{\"u}ller, Eckardt, Bach, Benz and Haydn}]{hofer2006enmap}
\bibinfo{author}{Hofer, S.}, \bibinfo{author}{Kaufmann, H.},
  \bibinfo{author}{Stuffler, T.}, \bibinfo{author}{Penn{\'e}, B.},
  \bibinfo{author}{Schreier, G.}, \bibinfo{author}{M{\"u}ller, A.},
  \bibinfo{author}{Eckardt, A.}, \bibinfo{author}{Bach, H.},
  \bibinfo{author}{Benz, U.}, \bibinfo{author}{Haydn, R.},
  \bibinfo{year}{2006}.
\newblock \bibinfo{title}{Enmap hyperspectral imager: an advanced optical
  payload for future applications in earth observation programs}, in:
  \bibinfo{booktitle}{Remote Sensing for Environmental Monitoring, GIS
  Applications, and Geology VI}, \bibinfo{organization}{SPIE}. pp.
  \bibinfo{pages}{94--99}.
\bibitem[{Hu et~al.(2019)Hu, Peng, Zhou, Xu, Zhao, Jiang, Fu, Wang and
  Shi}]{hu2019quantitative}
\bibinfo{author}{Hu, J.}, \bibinfo{author}{Peng, J.}, \bibinfo{author}{Zhou,
  Y.}, \bibinfo{author}{Xu, D.}, \bibinfo{author}{Zhao, R.},
  \bibinfo{author}{Jiang, Q.}, \bibinfo{author}{Fu, T.}, \bibinfo{author}{Wang,
  F.}, \bibinfo{author}{Shi, Z.}, \bibinfo{year}{2019}.
\newblock \bibinfo{title}{Quantitative estimation of soil salinity using
  uav-borne hyperspectral and satellite multispectral images}.
\newblock \bibinfo{journal}{Remote Sensing} \bibinfo{volume}{11},
  \bibinfo{pages}{736}.
\bibitem[{Khanal et~al.(2018)Khanal, Fulton, Klopfenstein, Douridas and
  Shearer}]{khanal2018integration}
\bibinfo{author}{Khanal, S.}, \bibinfo{author}{Fulton, J.},
  \bibinfo{author}{Klopfenstein, A.}, \bibinfo{author}{Douridas, N.},
  \bibinfo{author}{Shearer, S.}, \bibinfo{year}{2018}.
\newblock \bibinfo{title}{Integration of high resolution remotely sensed data
  and machine learning techniques for spatial prediction of soil properties and
  corn yield}.
\newblock \bibinfo{journal}{Computers and electronics in agriculture}
  \bibinfo{volume}{153}, \bibinfo{pages}{213--225}.
\bibitem[{Li et~al.(2020)Li, Yin, Cong and Du}]{li2020simultaneous}
\bibinfo{author}{Li, R.}, \bibinfo{author}{Yin, B.}, \bibinfo{author}{Cong,
  Y.}, \bibinfo{author}{Du, Z.}, \bibinfo{year}{2020}.
\newblock \bibinfo{title}{Simultaneous prediction of soil properties using
  multi\_cnn model}.
\newblock \bibinfo{journal}{Sensors} \bibinfo{volume}{20},
  \bibinfo{pages}{6271}.
\bibitem[{Loizzo et~al.(2018)Loizzo, Guarini, Longo, Scopa, Formaro,
  Facchinetti and Varacalli}]{loizzo2018prisma}
\bibinfo{author}{Loizzo, R.}, \bibinfo{author}{Guarini, R.},
  \bibinfo{author}{Longo, F.}, \bibinfo{author}{Scopa, T.},
  \bibinfo{author}{Formaro, R.}, \bibinfo{author}{Facchinetti, C.},
  \bibinfo{author}{Varacalli, G.}, \bibinfo{year}{2018}.
\newblock \bibinfo{title}{Prisma: The italian hyperspectral mission}, in:
  \bibinfo{booktitle}{IGARSS 2018-2018 IEEE International Geoscience and Remote
  Sensing Symposium}, \bibinfo{organization}{IEEE}. pp.
  \bibinfo{pages}{175--178}.
\bibitem[{Mammadov et~al.(2021)Mammadov, Nowosad and
  Glaesser}]{mammadov2021estimation}
\bibinfo{author}{Mammadov, E.}, \bibinfo{author}{Nowosad, J.},
  \bibinfo{author}{Glaesser, C.}, \bibinfo{year}{2021}.
\newblock \bibinfo{title}{Estimation and mapping of surface soil properties in
  the caucasus mountains, azerbaijan using high-resolution remote sensing
  data}.
\newblock \bibinfo{journal}{Geoderma Regional} \bibinfo{volume}{26},
  \bibinfo{pages}{e00411}.
\bibitem[{Mancini et~al.(2019)Mancini, Toscano and Rinnan}]{mancini2019study}
\bibinfo{author}{Mancini, M.}, \bibinfo{author}{Toscano, G.},
  \bibinfo{author}{Rinnan, {\AA}.}, \bibinfo{year}{2019}.
\newblock \bibinfo{title}{Study of the scattering effects on nir data for the
  prediction of ash content using emsc correction factors}.
\newblock \bibinfo{journal}{Journal of Chemometrics} \bibinfo{volume}{33},
  \bibinfo{pages}{e3111}.
\bibitem[{Meng et~al.(2020)Meng, Bao, Liu, Liu, Zhang, Zhang, Wang, Tang and
  Kong}]{meng}
\bibinfo{author}{Meng, X.}, \bibinfo{author}{Bao, Y.}, \bibinfo{author}{Liu,
  J.}, \bibinfo{author}{Liu, H.}, \bibinfo{author}{Zhang, X.},
  \bibinfo{author}{Zhang, Y.}, \bibinfo{author}{Wang, P.},
  \bibinfo{author}{Tang, H.}, \bibinfo{author}{Kong, F.}, \bibinfo{year}{2020}.
\newblock \bibinfo{title}{Regional soil organic carbon prediction model based
  on a discrete wavelet analysis of hyperspectral satellite data}.
\newblock \bibinfo{journal}{International Journal of Applied Earth Observation
  and Geoinformation} \bibinfo{volume}{89}, \bibinfo{pages}{102111}.
\newblock \DOIprefix\doi{10.1016/j.jag.2020.102111}.
\bibitem[{Menon and Seelamantula(2014)}]{menon2014robust}
\bibinfo{author}{Menon, S.V.}, \bibinfo{author}{Seelamantula, C.S.},
  \bibinfo{year}{2014}.
\newblock \bibinfo{title}{Robust savitzky-golay filters}, in:
  \bibinfo{booktitle}{2014 19th International Conference on Digital Signal
  Processing}, \bibinfo{organization}{IEEE}. pp. \bibinfo{pages}{688--693}.
\bibitem[{Mulder et~al.(2011)Mulder, de~Bruin, Schaepman and Mayr}]{mulder}
\bibinfo{author}{Mulder, V.}, \bibinfo{author}{de~Bruin, S.},
  \bibinfo{author}{Schaepman, M.}, \bibinfo{author}{Mayr, T.},
  \bibinfo{year}{2011}.
\newblock \bibinfo{title}{The use of remote sensing in soil and terrain mapping
  - a review}.
\newblock \bibinfo{journal}{Geoderma} \bibinfo{volume}{162},
  \bibinfo{pages}{1--19}.
\newblock \DOIprefix\doi{10.1016/j.geoderma.2010.12.018}.
\bibitem[{Nicolopoulou-Stamati et~al.(2016)Nicolopoulou-Stamati, Maipas,
  Kotampasi, Stamatis and Hens}]{nicolopoulou2016chemical}
\bibinfo{author}{Nicolopoulou-Stamati, P.}, \bibinfo{author}{Maipas, S.},
  \bibinfo{author}{Kotampasi, C.}, \bibinfo{author}{Stamatis, P.},
  \bibinfo{author}{Hens, L.}, \bibinfo{year}{2016}.
\newblock \bibinfo{title}{Chemical pesticides and human health: the urgent need
  for a new concept in agriculture}.
\newblock \bibinfo{journal}{Frontiers in public health} \bibinfo{volume}{4},
  \bibinfo{pages}{148}.
\bibitem[{Panagos et~al.(2012)Panagos, Van~Liedekerke, Jones and
  Montanarella}]{panagos2012european}
\bibinfo{author}{Panagos, P.}, \bibinfo{author}{Van~Liedekerke, M.},
  \bibinfo{author}{Jones, A.}, \bibinfo{author}{Montanarella, L.},
  \bibinfo{year}{2012}.
\newblock \bibinfo{title}{European soil data centre: Response to european
  policy support and public data requirements}.
\newblock \bibinfo{journal}{Land use policy} \bibinfo{volume}{29},
  \bibinfo{pages}{329--338}.
\bibitem[{Pierce and Nowak(1999)}]{pierce1999aspects}
\bibinfo{author}{Pierce, F.J.}, \bibinfo{author}{Nowak, P.},
  \bibinfo{year}{1999}.
\newblock \bibinfo{title}{Aspects of precision agriculture}.
\newblock \bibinfo{journal}{Advances in agronomy} \bibinfo{volume}{67},
  \bibinfo{pages}{1--85}.
\bibitem[{Ritter et~al.(2008)Ritter, Dicke, Weis, Oebel, Piepho, Büchse and
  Gerhards}]{ritter}
\bibinfo{author}{Ritter, C.}, \bibinfo{author}{Dicke, D.},
  \bibinfo{author}{Weis, M.}, \bibinfo{author}{Oebel, H.},
  \bibinfo{author}{Piepho, H.P.}, \bibinfo{author}{Büchse, A.},
  \bibinfo{author}{Gerhards, R.}, \bibinfo{year}{2008}.
\newblock \bibinfo{title}{An on-farm approach to quantify yield variation and
  to derive decision rules for site-specific weed management}.
\newblock \bibinfo{journal}{Precision Agriculture} \bibinfo{volume}{9},
  \bibinfo{pages}{133--146}.
\newblock \DOIprefix\doi{10.1007/s11119-008-9061-5}.
\bibitem[{Selvaraju et~al.(2017)Selvaraju, Cogswell, Das, Vedantam, Parikh and
  Batra}]{gradcam}
\bibinfo{author}{Selvaraju, R.R.}, \bibinfo{author}{Cogswell, M.},
  \bibinfo{author}{Das, A.}, \bibinfo{author}{Vedantam, R.},
  \bibinfo{author}{Parikh, D.}, \bibinfo{author}{Batra, D.},
  \bibinfo{year}{2017}.
\newblock \bibinfo{title}{Grad-cam: Visual explanations from deep networks via
  gradient-based localization}, in: \bibinfo{booktitle}{Proceedings of the IEEE
  international conference on computer vision}, pp. \bibinfo{pages}{618--626}.
\bibitem[{T{\'o}th et~al.(2013a)T{\'o}th, Jones and Montanarella}]{LUCAS}
\bibinfo{author}{T{\'o}th, G.}, \bibinfo{author}{Jones, A.},
  \bibinfo{author}{Montanarella, L.}, \bibinfo{year}{2013}a.
\newblock \bibinfo{title}{The lucas topsoil database and derived information on
  the regional variability of cropland topsoil properties in the european
  union}.
\newblock \bibinfo{journal}{Environmental monitoring and assessment}
  \bibinfo{volume}{185}.
\newblock \DOIprefix\doi{10.1007/s10661-013-3109-3}.
\bibitem[{T{\'o}th et~al.(2013b)T{\'o}th, Jones and
  Montanarella}]{toth2013LUCAS}
\bibinfo{author}{T{\'o}th, G.}, \bibinfo{author}{Jones, A.},
  \bibinfo{author}{Montanarella, L.}, \bibinfo{year}{2013}b.
\newblock \bibinfo{title}{LUCAS Topsoil Survey: methodology, data and results}.
\newblock \bibinfo{publisher}{Publications Office}.
\bibitem[{Vapnik(2013)}]{vapnik2013nature}
\bibinfo{author}{Vapnik, V.}, \bibinfo{year}{2013}.
\newblock \bibinfo{title}{The nature of statistical learning theory}.
\newblock \bibinfo{publisher}{Springer science \& business media}.
\bibitem[{Vohland et~al.(2017)Vohland, Ludwig, Thiele-Bruhn and
  Ludwig}]{vohland2017quantification}
\bibinfo{author}{Vohland, M.}, \bibinfo{author}{Ludwig, M.},
  \bibinfo{author}{Thiele-Bruhn, S.}, \bibinfo{author}{Ludwig, B.},
  \bibinfo{year}{2017}.
\newblock \bibinfo{title}{Quantification of soil properties with hyperspectral
  data: Selecting spectral variables with different methods to improve
  accuracies and analyze prediction mechanisms}.
\newblock \bibinfo{journal}{Remote Sensing} \bibinfo{volume}{9},
  \bibinfo{pages}{1103}.
\bibitem[{Wang et~al.(2022)Wang, Guan, Zhang, Lee, Margenot, Ge, Peng, Zhou,
  Zhou and Huang}]{wang2022using}
\bibinfo{author}{Wang, S.}, \bibinfo{author}{Guan, K.}, \bibinfo{author}{Zhang,
  C.}, \bibinfo{author}{Lee, D.}, \bibinfo{author}{Margenot, A.J.},
  \bibinfo{author}{Ge, Y.}, \bibinfo{author}{Peng, J.}, \bibinfo{author}{Zhou,
  W.}, \bibinfo{author}{Zhou, Q.}, \bibinfo{author}{Huang, Y.},
  \bibinfo{year}{2022}.
\newblock \bibinfo{title}{Using soil library hyperspectral reflectance and
  machine learning to predict soil organic carbon: Assessing potential of
  airborne and spaceborne optical soil sensing}.
\newblock \bibinfo{journal}{Remote Sensing of Environment}
  \bibinfo{volume}{271}, \bibinfo{pages}{112914}.
\bibitem[{Zhou et~al.(2021)Zhou, Geng, Ji, Xu, Wang, Pan, Bumberger, Haase and
  Lausch}]{zhou2021prediction}
\bibinfo{author}{Zhou, T.}, \bibinfo{author}{Geng, Y.}, \bibinfo{author}{Ji,
  C.}, \bibinfo{author}{Xu, X.}, \bibinfo{author}{Wang, H.},
  \bibinfo{author}{Pan, J.}, \bibinfo{author}{Bumberger, J.},
  \bibinfo{author}{Haase, D.}, \bibinfo{author}{Lausch, A.},
  \bibinfo{year}{2021}.
\newblock \bibinfo{title}{Prediction of soil organic carbon and the c: N ratio
  on a national scale using machine learning and satellite data: A comparison
  between sentinel-2, sentinel-3 and landsat-8 images}.
\newblock \bibinfo{journal}{Science of The Total Environment}
  \bibinfo{volume}{755}, \bibinfo{pages}{142661}.

\end{thebibliography}

\section*{Appendix}
In this section we report sensitivity results evaluated in terms of MAE, MSE, RMSE and Pearson score.
We also report the visual comparison between the proposed method and the state of the art for all the variables.
\newpage

\begin{table*}
\begin{adjustbox}{width=\textwidth}
\begin{tabular}{llrrrrrrrrrrrrr}
\toprule
 param. & value &  coarse &   clay &    silt &   sand &  $pH_{CaCl2}$ &  $pH_{H2O}$ &     OC &  CaCO3 &      N &       P &      K &    CEC &  global \\
 \midrule
 \multirow{3}{*}{fmin} &       450 &  0.1109 & 0.1565 &  0.1013 & 0.1044 &       0.1564 &     0.0816 & 0.1542 & 0.1076 & 0.0882 &  0.1448 & 0.1055 & 0.0833 &  0.1330 \\
     &       800 &  0.1095 & 0.1558 &  0.1003 & 0.0992 &       0.1526 &     0.1058 & 0.1580 & 0.1066 & 0.0918 &  0.1441 & 0.1054 & 0.0943 &  0.1345 \\
     &      1200 &  0.1123 & 0.1533 &  0.1070 & 0.0989 &       0.1499 &     0.1031 & 0.1610 & 0.1054 & 0.0882 &  0.1409 & 0.1037 & 0.0838 &  0.1347 \\
\midrule
\multirow{3}{*}{fmax} &      2500 &  0.1072 & 0.1535 &  0.1040 & 0.1003 &       0.1542 &     0.0958 & 0.1611 & 0.1096 & 0.0942 &  0.1455 &  0.1085 & 0.0847 &  0.1333 \\
     &      2300 &  0.1102 & 0.1561 &  0.1015 & 0.0986 &       0.1515 &     0.0982 & 0.1541 & 0.1054 & 0.0873 &  0.1413 &  0.1068 & 0.0888 &  0.1341 \\
     &      2400 &  0.1153 & 0.1561 &  0.1031 & 0.1036 &       0.1533 &     0.0966 & 0.1581 & 0.1045 & 0.0868 &  0.1430 &  0.0994 & 0.0879 &  0.1349 \\
\midrule
\multirow{3}{*}{insz} &       512 &  0.1137 & 0.1558 &  0.1040 & 0.0998 &       0.1529 &     0.0981 & 0.1602 & 0.1022 & 0.0883 &  0.1423 & 0.1052 & 0.0881 &  0.1335 \\
     &      1024 &  0.1073 & 0.1563 &  0.1046 & 0.0978 &       0.1527 &     0.0941 & 0.1581 & 0.1046 & 0.0885 &  0.1422 & 0.1066 & 0.0863 &  0.1337 \\
     &      2048 &  0.1118 & 0.1536 &  0.0999 & 0.1050 &       0.1534 &     0.0985 & 0.1549 & 0.1128 & 0.0915 &  0.1453 & 0.1029 & 0.0870 &  0.1350 \\
\midrule
\multirow{2}{*}{leak} &       0.2 &  0.1113 & 0.1519 &  0.1028 & 0.1010 &       0.1530 &     0.0958 & 0.1520 & 0.1003 & 0.0856 &  0.1445 &  0.1041 & 0.0859 &  0.1323 \\
     &       0.0 &  0.1105 & 0.1585 &  0.1029 & 0.1006 &       0.1530 &     0.0980 & 0.1634 & 0.1127 & 0.0932 &  0.1420 &  0.1056 & 0.0884 &  0.1358 \\
\midrule
\multirow{3}{*}{loss} &             l1 &  0.0976 &  0.1379 &  0.0923 &  0.0831 &       0.1434 &     0.0724 &  0.1573 & 0.0755 &  0.0633 &  0.1330 &  0.0914 &  0.0624 &  0.1179 \\
     &             l2 &  0.0986 &  0.1308 &  0.1003 &  0.0966 &       0.1514 &     0.0642 &  0.1522 & 0.0843 &  0.0668 &  0.1460 &  0.0948 &  0.0643 &  0.1202 \\
     & class. &  0.1364 &  0.1967 &  0.1159 &  0.1227 &       0.1641 &     0.1537 &  0.1637 & 0.1595 &  0.1380 &  0.1508 &  0.1284 &  0.1345 &  0.1640 \\
     \midrule
\multirow{2}{*}{b.n.} &      True &  0.0977 &  0.1117 &  0.0704 &  0.0646 &       0.0697 &     0.0981 &  0.1026 &  0.0953 &  0.0839 &  0.0845 &  0.0735 &  0.0817 &  0.0877 \\
     &     False &  0.1242 &  0.1990 &  0.1355 &  0.1373 &       0.2368 &     0.0956 &  0.2133 &  0.1177 &  0.0950 &  0.2024 &  0.1365 &  0.0926 &  0.1807 \\
     
\bottomrule
\end{tabular}

\end{adjustbox}
\caption{Sensitivity analysis performed with MAE score.}
\label{fig:san_mae}
\end{table*}

\begin{table*}
\begin{adjustbox}{width=\textwidth}
\begin{tabular}{llrrrrrrrrrrrrr}
\toprule
 param. & value &  coarse &   clay &    silt &   sand &  $pH_{CaCl2}$ &  $pH_{H2O}$ &     OC &  CaCO3 &      N &       P &      K &    CEC &  global \\ %
 
 \midrule
\multirow{3}{*}{fmin} &       450 &  0.0463 & 0.0832 &  0.0460 & 0.0449 &       0.1015 &     0.0269 & 0.0814 & 0.0609 & 0.0365 &  0.0707 & 0.0458 & 0.0311 &  0.0940 \\
                      &      1200 &  0.0497 & 0.0766 &  0.0432 & 0.0440 &       0.0918 &     0.0464 & 0.0819 & 0.0548 & 0.0373 &  0.0637 & 0.0403 & 0.0331 &  0.0959 \\
                      &       800 &  0.0459 & 0.0814 &  0.0446 & 0.0433 &       0.0981 &     0.0484 & 0.0842 & 0.0646 & 0.0416 &  0.0688 & 0.0439 & 0.0394 &  0.0964 \\
\midrule
\multirow{3}{*}{fmax} &      2300 &  0.0439 & 0.0799 &  0.0419 & 0.0415 &       0.0943 &     0.0421 & 0.0772 & 0.0569 & 0.0354 &  0.0649 &  0.0430 & 0.0339 &  0.0939 \\
                      &      2500 &  0.0463 & 0.0792 &  0.0452 & 0.0446 &       0.0980 &     0.0395 & 0.0867 & 0.0642 & 0.0442 &  0.0690 &  0.0461 & 0.0348 &  0.0951 \\
                      &      2400 &  0.0518 & 0.0820 &  0.0467 & 0.0461 &       0.0989 &     0.0402 & 0.0836 & 0.0593 & 0.0358 &  0.0692 &  0.0409 & 0.0349 &  0.0973 \\
\midrule
\multirow{3}{*}{insz} &      1024 &  0.0442 & 0.0815 &  0.0454 & 0.0412 &       0.0974 &     0.0375 & 0.0838 & 0.0559 & 0.0360 &  0.0675 & 0.0446 & 0.0336 &  0.0946 \\
                      &       512 &  0.0484 & 0.0816 &  0.0466 & 0.0432 &       0.0990 &     0.0407 & 0.0855 & 0.0549 & 0.0367 &  0.0689 & 0.0444 & 0.0346 &  0.0950 \\
                      &      2048 &  0.0493 & 0.0780 &  0.0418 & 0.0478 &       0.0949 &     0.0436 & 0.0781 & 0.0696 & 0.0426 &  0.0668 & 0.0410 & 0.0354 &  0.0967 \\
\midrule
\multirow{2}{*}{leak} &       0.2 &  0.0483 & 0.0763 &  0.0439 & 0.0447 &       0.0966 &     0.0391 & 0.0778 & 0.0523 & 0.0349 &  0.0686 &  0.0419 & 0.0327 &  0.0931 \\
                      &       0.0 &  0.0463 & 0.0844 &  0.0453 & 0.0434 &       0.0975 &     0.0421 & 0.0872 & 0.0679 & 0.0420 &  0.0668 &  0.0447 & 0.0364 &  0.0977 \\
\midrule
\multirow{3}{*}{loss} &     l2 &  0.0309 &  0.0522 &  0.0370 &  0.0303 &       0.0905 &     0.0123 &  0.0696 & 0.0272 &  0.0164 &  0.0625 &  0.0305 &  0.0127 &  0.0736 \\
                      &     l1 &  0.0333 &  0.0610 &  0.0330 &  0.0261 &       0.0847 &     0.0191 &  0.0746 & 0.0226 &  0.0160 &  0.0539 &  0.0287 &  0.0145 &  0.0752 \\
                      & class. &  0.0776 &  0.1277 &  0.0638 &  0.0756 &       0.1161 &     0.0902 &  0.1032 & 0.1303 &  0.0828 &  0.0867 &  0.0706 &  0.0762 &  0.1373 \\
\midrule
\multirow{2}{*}{b.n.} &      True &  0.0359 &  0.0404 &  0.0140 &  0.0186 &       0.0194 &     0.0474 &  0.0288 &  0.0625 &  0.0368 &  0.0198 &  0.0219 &  0.0318 &  0.0409 \\
                      &     False &  0.0587 &  0.1206 &  0.0754 &  0.0696 &       0.1753 &     0.0338 &  0.1366 &  0.0577 &  0.0402 &  0.1160 &  0.0649 &  0.0373 &  0.1503 \\
     
\bottomrule
\end{tabular}

\end{adjustbox}
\caption{Sensitivity analysis performed with MSE score.}
\label{fig:san_mse}
\end{table*}

\begin{table*}
\begin{adjustbox}{width=\textwidth}
\begin{tabular}{llrrrrrrrrrrrrr}
\toprule
 param. & value &  coarse &   clay &    silt &   sand &  $pH_{CaCl2}$ &  $pH_{H2O}$ &     OC &  CaCO3 &      N &       P &      K &    CEC &  global \\
 \midrule
 
\multirow{3}{*}{fmin} &       450 &  0.2366 & 0.5476 &  0.4359 & 0.5198 &       0.6616 &     0.6571 & 0.5401 & 0.5410 & 0.4920 &  0.2310 & 0.3595 & 0.5637 &  0.5166 \\
                      &       800 &  0.2335 & 0.4933 &  0.4085 & 0.4891 &       0.5797 &     0.6060 & 0.4823 & 0.4851 & 0.4801 &  0.1985 & 0.3088 & 0.5199 &  0.4838 \\
                      &      1200 &  0.2115 & 0.5005 &  0.3814 & 0.4542 &       0.5836 &     0.6041 & 0.4902 & 0.5077 & 0.4808 &  0.1878 & 0.3042 & 0.4790 &  0.4720 \\
\midrule
\multirow{3}{*}{fmax} &      2500 &  0.2290 & 0.5509 &  0.4112 & 0.4985 &       0.6278 &     0.6323 & 0.5207 & 0.5372 & 0.4895 &  0.2062 &  0.3176 & 0.5493 &  0.4950 \\
                      &      2400 &  0.2345 & 0.5283 &  0.4289 & 0.4861 &       0.6231 &     0.6241 & 0.5171 & 0.5267 & 0.4912 &  0.2087 &  0.3360 & 0.5115 &  0.4931 \\
                      &      2300 &  0.2180 & 0.4660 &  0.3839 & 0.4762 &       0.5725 &     0.6085 & 0.4731 & 0.4698 & 0.4725 &  0.2016 &  0.3156 & 0.4983 &  0.4830 \\
\midrule
\multirow{3}{*}{insz} &      2048 &  0.2455 & 0.5960 &  0.4435 & 0.5239 &       0.6704 &     0.6743 & 0.5498 & 0.5498 & 0.5146 &  0.2279 & 0.3483 & 0.5654 &  0.5493 \\
                      &      1024 &  0.2377 & 0.5017 &  0.3893 & 0.4719 &       0.5957 &     0.6383 & 0.5094 & 0.5063 & 0.4946 &  0.2028 & 0.3128 & 0.5084 &  0.4941 \\
                      &       512 &  0.2027 & 0.4579 &  0.3951 & 0.4699 &       0.5655 &     0.5632 & 0.4588 & 0.4808 & 0.4496 &  0.1898 & 0.3112 & 0.4917 &  0.4414 \\
\midrule
\multirow{2}{*}{leak} &       0.0 &  0.2378 & 0.5154 &  0.3881 & 0.4808 &       0.6177 &     0.6320 & 0.5142 & 0.5318 & 0.4742 &  0.2056 &  0.3266 & 0.5196 &  0.4963 \\
                      &       0.2 &  0.2173 & 0.5114 &  0.4262 & 0.4924 &       0.5965 &     0.6114 & 0.4927 & 0.4900 & 0.4939 &  0.2052 &  0.3196 & 0.5191 &  0.4848 \\
\midrule
\multirow{3}{*}{loss} &             l2 &  0.4429 &  0.8049 &  0.6427 &  0.7320 &       0.9353 &     0.9331 &  0.8398 & 0.7095 &  0.8608 &  0.3242 &  0.4856 &  0.8221 &  0.7111 \\
                      &             l1 &  0.2688 &  0.5329 &  0.4866 &  0.5184 &       0.5807 &     0.5784 &  0.5428 & 0.5900 &  0.5584 &  0.3225 &  0.3400 &  0.5404 &  0.5135 \\
                      & class. &  0.0576 &  0.3218 &  0.1877 &  0.3080 &       0.4367 &     0.4741 &  0.2656 & 0.3297 &  0.2024 &  0.0332 &  0.2143 &  0.3174 &  0.3112 \\
\midrule
\multirow{2}{*}{b.n.} &      True &  0.3246 &  0.7540 &  0.5612 &  0.6802 &       0.8703 &     0.8718 &  0.7318 &  0.7315 &  0.6962 &  0.2887 &  0.4617 &  0.7283 &  0.6417 \\
                      &     False & -0.0054 & -0.0516 &  0.0183 &  0.0123 &      -0.0115 &     0.0247 & -0.0340 &  0.0263 & -0.0136 &  0.0039 &  0.0018 &  0.0288 & -0.0000 \\
     
\bottomrule
\end{tabular}

\end{adjustbox}
\caption{Sensitivity analysis performed with Pearson score.}
\label{fig:san_pearson}
\end{table*}

\begin{table*}
\begin{adjustbox}{width=\textwidth}
\begin{tabular}{llrrrrrrrrrrrrr}
\toprule
 param. & value &  coarse &   clay &    silt &   sand &  $pH_{CaCl2}$ &  $pH_{H2O}$ &     OC &  CaCO3 &      N &       P &      K &    CEC &  global \\
 \midrule
\multirow{3}{*}{fmin} &       450 &  0.1989 & 0.2629 &  0.1844 & 0.1912 &       0.2749 &     0.1345 & 0.2493 & 0.2067 & 0.1665 &  0.2329 & 0.1857 & 0.1492 &  0.2839 \\
                      &      1200 &  0.2026 & 0.2551 &  0.1891 & 0.1802 &       0.2625 &     0.1812 & 0.2589 & 0.2017 & 0.1642 &  0.2248 & 0.1825 & 0.1533 &  0.2900 \\
                      &       800 &  0.1981 & 0.2621 &  0.1848 & 0.1841 &       0.2719 &     0.1880 & 0.2573 & 0.2096 & 0.1758 &  0.2324 & 0.1870 & 0.1741 &  0.2904 \\
\midrule
\multirow{3}{*}{fmax} &      2500 &  0.1926 & 0.2546 &  0.1874 & 0.1838 &       0.2718 &     0.1649 & 0.2616 & 0.2144 & 0.1784 &  0.2338 &  0.1915 & 0.1546 &  0.2862 \\
                      &      2300 &  0.1974 & 0.2629 &  0.1822 & 0.1817 &       0.2664 &     0.1721 & 0.2480 & 0.2025 & 0.1640 &  0.2258 &  0.1892 & 0.1622 &  0.2876 \\
                      &      2400 &  0.2095 & 0.2626 &  0.1889 & 0.1900 &       0.2710 &     0.1669 & 0.2561 & 0.2011 & 0.1642 &  0.2305 &  0.1745 & 0.1600 &  0.2906 \\
\midrule
\multirow{3}{*}{insz} &       512 &  0.2045 & 0.2612 &  0.1891 & 0.1828 &       0.2702 &     0.1685 & 0.2620 & 0.1971 & 0.1662 &  0.2295 & 0.1886 & 0.1595 &  0.2865 \\
                      &      1024 &  0.1923 & 0.2607 &  0.1886 & 0.1795 &       0.2695 &     0.1635 & 0.2559 & 0.2000 & 0.1647 &  0.2291 & 0.1862 & 0.1572 &  0.2868 \\
                      &      2048 &  0.2029 & 0.2582 &  0.1806 & 0.1932 &       0.2695 &     0.1720 & 0.2477 & 0.2210 & 0.1756 &  0.2314 & 0.1803 & 0.1601 &  0.2911 \\
\midrule
\multirow{2}{*}{leak} &       0.2 &  0.2011 & 0.2523 &  0.1853 & 0.1868 &       0.2697 &     0.1663 & 0.2441 & 0.1909 & 0.1601 &  0.2332 &  0.1825 & 0.1555 &  0.2833 \\
                      &       0.0 &  0.1987 & 0.2678 &  0.1869 & 0.1835 &       0.2698 &     0.1697 & 0.2663 & 0.2210 & 0.1774 &  0.2268 &  0.1876 & 0.1623 &  0.2929 \\
\midrule
\multirow{3}{*}{loss} &             l2 &  0.1696 &  0.2091 &  0.1733 &  0.1615 &       0.2613 &     0.0979 &  0.2408 & 0.1538 &  0.1199 &  0.2254 &  0.1649 &  0.1069 &  0.2505 \\
                      &             l1 &  0.1773 &  0.2290 &  0.1655 &  0.1457 &       0.2505 &     0.1184 &  0.2480 & 0.1354 &  0.1197 &  0.2054 &  0.1566 &  0.1116 &  0.2540 \\
                      & class. &  0.2526 &  0.3415 &  0.2195 &  0.2480 &       0.2974 &     0.2869 &  0.2768 & 0.3283 &  0.2665 &  0.2591 &  0.2335 &  0.2578 &  0.3596 \\
\midrule
\multirow{2}{*}{b.n.} &      True &  0.1758 &  0.1815 &  0.1113 &  0.1216 &       0.1311 &     0.1836 &  0.1524 &  0.1869 &  0.1544 &  0.1299 &  0.1296 &  0.1499 &  0.1923 \\
                      &     False &  0.2242 &  0.3391 &  0.2614 &  0.2491 &       0.4093 &     0.1523 &  0.3587 &  0.2252 &  0.1833 &  0.3307 &  0.2408 &  0.1680 &  0.3846 \\

\bottomrule
\end{tabular}

\end{adjustbox}
\caption{Sensitivity analysis performed with RMSE score.}
\label{fig:san_rmse}
\end{table*}

\mapbig{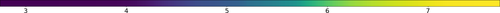}{$pH_{CaCl2}$}
\mapbig{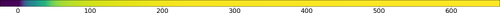}{$CaCO3$}
\mapbig{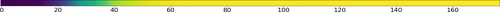}{$OC$}
\mapbig{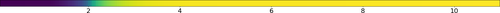}{$N$}
\mapbig{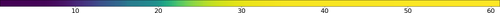}{$clay$}
\mapbig{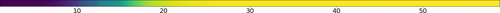}{$CEC$}
\mapbig{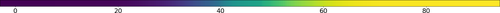}{$sand$}
\mapbig{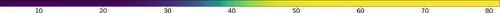}{$silt$}
\mapbig{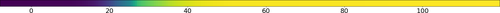}{$P$}
\mapbig{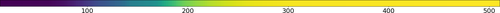}{$K$}
\mapbig{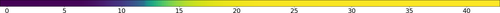}{$coarse$}

\end{document}